\documentclass[sigconf]{acmart}

\newcommand{\ourmodel}{ISRR }
\newcommand{\by}{\mathbf{y} }

\AtBeginDocument{%
  \providecommand\BibTeX{{%
    \normalfont B\kern-0.5em{\scshape i\kern-0.25em b}\kern-0.8em\TeX}}}

\setcopyright{acmcopyright}
\copyrightyear{2023}
\acmYear{2023}
\acmDOI{XXXXXXX.XXXXXXX}

\acmPrice{15.00}
\acmISBN{978-1-4503-XXXX-X/18/06}

\usepackage{bm}
\usepackage{dsfont}
\usepackage{subcaption}
\usepackage{enumitem}
\usepackage{amsmath}
\usepackage{titlesec}

\usepackage{caption}

\usepackage{enumitem}
 \usepackage{multirow}
 \usepackage{multicol}

\usepackage{amsmath}
\usepackage{algorithm}
\usepackage{algpseudocode}

\begin{document}

\title{Inference-time Stochastic Ranking with Risk Control}

\newcommand{\lgb}{LightGBM}
\newcommand{\cat}{CatBoost}
\newcommand{\yl}[1]{\textbf{\color{purple}(Yang: #1)}}
\theoremstyle{definition}

\newtheorem{assumption}{Assumption}

\title{Inference-time Stochastic Ranking with Risk Control}

\author{Ruocheng Guo}
\affiliation{%
  \institution{ByteDance Research}
  \country{UK}
}
\email{ruocheng.guo@bytedance.com}
\author{Jean-Fran\c cois Ton}
\affiliation{%
  \institution{ByteDance Research}
  \country{UK}
}
\email{jeanfrancois@bytedance.com}

\author{Yang Liu}
\affiliation{%
  \institution{ByteDance Research}
  \country{USA}
}
\email{yang.liu01@bytedance.com}

\author{Hang Li}
\affiliation{%
  \institution{ByteDance Research}
  \country{China}
}
\email{lihang.lh@bytedance.com}
\begin{abstract}

Learning to Rank (LTR) methods are vital in online economies, affecting users and item providers. Fairness in LTR models is crucial to allocate exposure proportionally to item relevance. Widely used deterministic LTR models can lead to unfair exposure distribution, especially when items with the same relevance receive slightly different ranking scores.
Stochastic LTR models, incorporating the Plackett-Luce (PL) ranking model, address fairness issues but suffer from high training cost. In addition, they cannot provide guarantees on the utility or fairness, which can lead to dramatic degraded utility when optimized for fairness.
To overcome these limitations, we propose Inference-time Stochastic Ranking with Risk Control (ISRR), a novel method that performs stochastic ranking at inference time with guanranteed utility or fairness given pretrained scoring functions from deterministic or stochastic LTR models.
Comprehensive experimental results on three widely adopted datasets demonstrate that our proposed method achieves utility and fairness comparable to existing stochastic ranking methods with much lower computational cost.
In addition, results verify that our method provides finite-sample guarantee on utility and fairness.
This advancement represents a significant contribution to the field of stochastic ranking and fair LTR with promising real-world applications.
\end{abstract}

\maketitle

















\section{Introduction}

In learning to rank (LTR), machine learning models are trained to optimize rankings of items for crucial applications in information retrieval \cite{schutze2008introduction,chapelle2011yahoo,liu2009learning} including e-commerce~\cite{guo2020debiasing}, streaming services~\cite{weston2013learning}, and sharing economy platforms~\cite{haldar2023learning}.
LTR models are playing a vital role in online economies involving not only users, but also item providers (e.g., sellers in e-commerce platforms and content creators in streaming websites), where ranking determines the exposure of items and has a substantial impact on the economic outcomes of sellers, job candidates, and content creators \cite{ ramanath2018towards,pei2019personalized} etc. 

An LTR model is typically composed of two components: a scoring function and a ranking model. Given a user query and a set of candidate items to be recommended, the scoring function predicts ranking scores for each item based on the predicted relevance to the query. Then, the ranking model generates a ranking list of products using the ranking scores.
Most traditional LTR models employ deterministic ranking models, which sort items in accordance to their ranking scores.

With the growing impact of LTR on online platforms, fair allocation of exposure among items~\cite{singh2019policy,morik2020controlling,oosterhuis2021computationally,diaz2020evaluating} becomes crucial. Fairness in ranking helps earn users' trust and promotes a healthy platform for item providers in the long run.
%
%
Considering the notion of amortized individual fairness for ranking that requires item exposure to be proportional to item relevance, people find that the widely adopted deterministic rankers tend to exhibit biases in the distribution of exposure~\cite{biega2018equity,singh2019policy,yang2023vertical}.
For instance, when two products with identical ground truth relevance have slightly different ranking scores, deterministic LTR models always rank the item with higher ranking score at a higher position, leading to unfair allocation of exposure~\cite{singh2019policy}.
%

To address this issue, there has been a shift towards stochastic LTR models. The representive methods~\cite{oosterhuis2021computationally,ustimenko2020stochasticrank,bruch2020stochastic,singh2019policy} incorporate the Plackett-Luce (PL) ranking model \cite{plackett1975analysis} which can predict a distribution over rankings of items based on their ranking scores. Given the same query, the PL model can sample a diverse set of ranking lists to improve exposure fairness~\cite{singh2019policy}.


However, the state-of-the-art stochastic LTR models need computationally intensive training to achieve high expected utility or fairness for rankings sampled from the PL model. In particular, it requires training on a large number of sampled rankings from the PL model for each input query~\cite{oosterhuis2021computationally,oosterhuis2022learning}.
In addition, incorporating scoring functions from pre-trained deterministic LTR models with the PL ranking model, although is computationally efficient, can result in subpar utility fairness trade-off.
This is because scoring functions from deterministic LTR models are not trained to optimize the expected utility or fairness under the ranking distributions predicted by the PL ranking model.
%
%
In practice, two major obstacles hinder the widespread adoption of stochastic LTR models in real-world applications: (1) the high training costs, as detailed in the experimental results in Section~\ref{subsubsec:running_time}, and (2) the absence of guarantees in either utility or fairness when considering the utility-fairness trade-off. 

To address these challenges, we present \textit{\underline{I}nference-time \underline{S}tochastic \underline{Rank}ing with \underline{R}isk Control} (ISRR), an inference-time ranking method for efficient and effective utility-fairness trade-off. 
This method represents the inaugural approach to implementing stochastic ranking at inference time with guaranteed utility or fairness. This property is maintained irrespective of whether the scoring function was pre-trained with deterministic or stochastic ranking models.
In particular, our proposed method consists of two components. First, we propose a generalized PL (GPL) ranking model.
GPL builds the foundation for an effective sampling of rankings based on the pre-trained scoring function of deterministic or stochastic LTR models, which allows us to circumvent the expensive training of stochastic LTR models.
Second, with a novel distribution-free risk control approach tailored for the GPL model, our method provides principled finite-sample guarantees, assuring a specified utility or fairness level.
The contributions of this work can be summarized as follows:
\begin{itemize}
[itemsep=0.1pt,topsep=0pt,leftmargin=*]
\item  We propose ISRR, an inference-time, model-agnostic method that creates a principled stochastic LTR model from the pre-trained scoring function of any LTR model.
It can enjoy the (1) low training cost of deterministic LTR models by directly adopting their pre-trained scoring function, (2) improved fairness of stochastic LTR models, and (3) finite-sample guarantee on fairness or utility.
%
\item In three benchmark datasets, extensive experimental results show that the proposed GPL model provides an effective utility fairness trade-off and subsumes the deterministic and the stochastic ranking model as special cases.
In addition, results show that ISRR can (a) guarantee specified levels of utility while (b) enhancing exposure-based fairness with (c) much lower running time than existing stochastic LTR models.
%
\end{itemize}




\section{Preliminaries}

In this section, we begin by outlining the notation used throughout the paper. Next, we formally define an LTR model, which consists of a scoring function and a ranking model. Following that, we introduce definitions for utility and amortized indivudal fairness measures within the context of Learning to Rank (LTR). Lastly, we conclude this section by presenting our problem statement.

\noindent\textbf{Notations.}
For a query $q$, there exists $n_q$ candidate documents $\mathcal{D}^q = \{d_1^q,...,d_{n_q}^q\}$ to be ranked. Each document $d_i^q$ is described by a tuple $(\mathbf{x}_i^q,\rho(d_i^q))$, where the feature vector $\mathbf{x}_i^q \in \mathcal{X}$ describes the item and its relationship to the query $q$. For example, features used in e-commerce search can include the price of the item and the average price of items clicked from the query. 
And $\rho(d_i^q)$ is the relevance of document $d_i^q$ annotated by human experts.
We assume that the relevance is given for each item corresponding to the queries in the training, validation and calibration set, but unknown for the test set.
For simplicity of notation, we will omit the subscript $i$ and the superscript $q$ when they are not necessary. 
A top-K ranking $\mathbf{y}=[y_1,...,y_K] \in \mathcal{Y}$ is a sorted list of $K$ items, where $y_k=d$ means item $d$ is ranked at the $k$-th position in $\mathbf{y}$, where $\mathcal{Y}$ is the space of top-k ranking lists.
Let $\mathbf{y}_{1:k}$ be the sublist including first $k\le K$ elements of $\mathbf{y}$.

\noindent\textbf{Scoring Function and Ranking Model.} An LTR model consists of two components: scoring function and ranking model.
Here, we formally define the scoring function and the ranking model of an LTR model.
First, given query $q$ and its item set $\mathcal{D}^q$, a scoring function $f:\mathcal{X} \rightarrow \mathbb{R}$ maps the feature vectors of each item $d$ to its ranking scores $s(d)$.
In this work, we assume that the scoring function $f$ is fixed and ranking scores $s(d)$ are given for $d \in \mathcal{D}^q$.
Second, a ranking model $\pi : \mathbb{R}^{n_q} \times \mathcal{Y} \rightarrow [0,1]$ which maps the scores of all items in $\mathcal{D}^q$ and a ranking $\by$ to its probability to be sampled. 
%
%
Thus, a ranking model $\pi(\{s(1),...,s(n_q)\},\by)$ predicts a distribution of rankings for each query $q$. For simplicity of notations, we denote the predicted distribution of rankings for query $q$ as $\pi^q(\by)$.
Then, a deterministic LTR model comes with $\pi^q(\mathbf{y})=1$ for a certain ranking $\mathbf{y}$, and $\pi^q(\mathbf{y}')=0$ for $\mathbf{y}'\ne \mathbf{y}$. While a stochastic model can have $\pi^q(\mathbf{y}) > 0$ for multiple different rankings $\mathbf{y}$. 

%
The PL ranking model is used in most existing stochastic LTR models~\citep{singh2019policy,diaz2020evaluating,oosterhuis2021computationally} .
%
%
PL ranking models predict a distribution of rankings for fairer allocation of exposure among items. 
%

%
Given query $q$ and the set of items $\mathcal{D}^q$, their ranking scores $s(d),d\in\mathcal{D}^q$, and the sampled items for positions $1,...,k-1$, denoted by  $\by_{1:k-1}$, the PL 
ranking model~\citep{plackett1975analysis} samples an item $d$ for position $k$ from $\pi_{PL}(\by) = \prod_{k=1}^K p_{PL}(d|\by_{1:k-1}) $ with:
\begin{equation}
\begin{split}
      p_{PL}(d|\by_{1:k-1}) = \frac{\mathds{1}(d\not \in \by_{1:k-1})\exp(s(d)/\tau)}{\sum_{d'\in \mathcal{D}^q\setminus \by_{1:k-1}}\exp(s(d')/\tau)}, 
\end{split}
\label{eq:pl_rank}
\end{equation}
where $\mathds{1}(d\not \in \by_{1:k-1})$ in the numerator implies that the sampling probability $p_{PL}(d|\by_{1:k-1})$ is nonzero iff $d$ has not been sampled before position $k$. Parameter $\tau$ is the temperature for the softmax.
%
%
However, training such stochastic LTR models is expensive as it requires sampling at least $100$ ranking lists for each query as in~\citep{oosterhuis2021computationally,bruch2020stochastic,singh2019policy} to accurately approximate gradients of the model parameters. 

\noindent\textbf{Ranking Metrics for Utility and Fairness.}
Given the definitions above, here, we define the utility and amortized indivudal fairness for an LTR model.
In LTR, the utility function considers the ranking of each item by weighting each position $k$ with weight $\theta_k$.
The utility of a ranking model $\pi$ on query $q$ can be defined as~\cite{oosterhuis2021computationally}:
\begin{equation*}
\setlength{\baselineskip}{0pt}
    U^q(\pi) = \sum_{\mathbf{y} \in  \mathcal{Y}} \pi^q(\mathbf{y}) \sum_{k=1}^K \theta_k \cdot \rho(y_k),
    \label{eq:util}
\end{equation*}
which leads to the overall utility $U(\pi) = \mathbb{E}_{q}[U^q(\pi)]$.
If we choose $\theta_k=\frac{\mathds{1}[k\le K]}{\log_2(1+k)}$, then $U(\pi|q)$ is DCG@K.
Let IDCG@k 
be the maximal DCG@k for a given query $q$ at position $k$, then $U^q(\pi)$ is NDCG@K if $\theta_k = \frac{\mathds{1}[k\le K]}{\log_2(1+k) \times \text{IDCG@k}}$, which measures the normalized exposure of items ranked at position $k$.
In this work, we consider bounded utility function $U^q(\pi)\in [0,1]$. Thus, the utility risk to be controlled is $R_{util}(\pi) = 1-U(\pi)$, e.g., $1-\text{NDCG@K}$.

Fairness in ranking deals with the allocation of exposure over items.
Exposure measures the probability of users to examine a certain position.
The widely used utility metric NDCG@K is based on the logarithmic reduction of exposure proportional to the position. 
To measure item exposure fairness, we first define exposure of item $d$ under the ranking model $\pi$ as
\[
\mathcal{E}^q(d;\pi) = \sum_{\mathbf{y}} \pi^q(\mathbf{y}) \sum_{k=1}^K \theta_k \cdot \mathds{1}[y_k=d],
\]
where $\mathds{1}[y_k=d]\theta_k$ is the exposure of item $d$ in the ranking $\mathbf{y}$.
Intuitively, it measures the mean exposure of item $d$ in the rankings sampled from the predicted 
 distribution $\pi^q(\mathbf{y})$.
 Let $\mathcal{E}(d)$ denote $\mathcal{E}^q(d;\pi)$ when $q$ and $\pi$ can be dropped.
 Based on this, we can define a disparity measure for amortized indivudal fairness in ranking.

 \noindent\textit{Amortized indivudal Fairness in Ranking~\citep{biega2018equity,singh2019policy,oosterhuis2021computationally}.}
%
In this work, we focus on fair allocation of exposures to items.
Singh and Joachims~\citep{singh2019policy} first propose that the exposure of an item $\mathcal{E}(d)$ should be proportional to its relevance $\rho(d)$.
%
%
They use the average difference of the exposure-relevance ratio $\frac{\mathcal{E}(d)}{\rho(d)} - \frac{\mathcal{E}(d')}{\rho(d')}$ between each pair of items for every query as the disparity measure.
%
Oosterhuis~\citep{oosterhuis2021computationally} proposes a variant of it, which handles the cases for the items with $0$ relevance but was ranked in top-K.
 Given a ranking model $\pi$, the disparity measure for amortized indivudal fairness is~\citep{oosterhuis2021computationally, singh2019policy}
\begin{equation*}
       R_{fair}^q(\pi) = \frac{2\sum_{d\in\mathcal{D}^q} \sum_{d'\in \mathcal{D}_{\neg d}^q }\ell(\mathcal{E}^q(d;\pi)\rho(d'),\mathcal{E}^q(d';\pi)\rho(d))}{|\mathcal{D}^q|(|\mathcal{D}^q|-1)},
       \label{eq:def_fair}
\end{equation*}
 where $\mathcal{D}_{\neg d}^q$ denotes $\mathcal{D}^q \setminus \{d\}$ and  $\ell(a,b)$ is $(a-b)^2$.
Let $R_{fair}(\pi)
=\mathbb{E}_q[R_{fair}^q(\pi)]$ be the expectation over queries.
Intuitively, $R_{fair}^q(\pi)$ measures how is the exposure of items under the ranking model $\pi$ different from the ideal case where the exposure is proportional to the relevance, $\frac{\mathcal{E}(d)}{\rho(d)} = \frac{\mathcal{E}(d')}{\rho(d')}$, for all pairs of items $d,d'\in\mathcal{D}^q$. 
%


\noindent\textbf{Problem Statement.}
%
%
For a given pre-trained scoring function $f$, we aim to maintain a specified level of utility (fairness) with high probability while optimizing exposure fairness (utility) of the LTR model via an inference-time method.
%

Formally, given query $q$ and the candidate items $\mathcal{D}^q$, and a fixed scoring function $f$, the goal is to optimize the ranking model $\pi$ to minimize the one risk function $R'$ (e.g., disparity) with a constraint that the other risk $R$ (e.g., utility risk) is bounded by a user specified level $\alpha$ with probability at least $1-\delta$.
This can be written as
\begin{equation}
   \underset{\pi}{\min} \; R'(\pi) \; s.t. \;P(R(\pi) \le \alpha) \ge 1-\delta, 
   \label{eq:prob_state}
\end{equation}
where $\alpha \in (0,1)$ ($1-\alpha$) is the desired level for $R$, and $1-\delta \in (0,1)$ is the desired coverage rate.
\section{Methodology}
\label{sec:method}

%
%
In this section, the background of distribution-free risk control is introduced, followed by the two components of the ISRR framework: the generalized PL ranking model and its risk control algorithm.
%



%


%

\subsection{Background: Distribution-free Risk Control}

Distribution-free risk control is a family of inference-time methods to offer finite-sample guarantees that a given risk function iis highly probable to remain at or below a user specified level.
In our setting, the proposed method uses a calibration set $\mathcal{Q}_{cal}$ to search for (some of) the model paramteres s.t. the constraint in Eq.~\eqref{eq:prob_state} is satisfied.

For simplicity, let $\mathcal{T}({\lambda})$ be a set-valued function with a scalar parameter $\lambda$ (e.g., a threshold on item scores) that predicts a set of items. Given a bounded risk function $R(\mathcal{T}(\lambda)) \in [0,B]$
that measures the expected loss of $\mathcal{T}(\lambda)$ over queries, we define of a Risk Controlling Prediction Set~\citep{bates2021distribution}. For simplicity, let $R(\lambda)$ denote $R(\mathcal{T}(\lambda))$.
%
%
Distribution-free risk control~\citep{bates2021distribution} uses an independent and identically distributed (i.i.d.) data split as the calibration set $\mathcal{Q}_{cal}$ to select $\lambda$ for the set-valued functions $\mathcal{T}(\lambda)$ s.t. the risk function $R$ is guaranteed on the test set $\mathcal{Q}_{test}$.
In our setting, set-valued functions predict a set of items for each position in the ranking. The connection between set-valued functions and ranking models is shown in Section~\ref{subsec:TPL}.
%

\noindent\textbf{Upper Confidence Bound (UCB) based Risk Control~\citep{bates2021distribution}.} We first define risk-controlling prediction sets.
    Given desired risk level $\alpha \in [0,B]$ and tolerance rate $\delta \in (0,1)$, a set-valued function $\mathcal{T}(\lambda)$ is a $(\alpha,\delta)$ risk-controlling prediction set iff
    \setlength{\abovedisplayskip}{2pt}
\setlength{\belowdisplayskip}{2pt}
\setlength{\abovedisplayshortskip}{2pt}
\setlength{\belowdisplayshortskip}{2pt}
    \begin{equation}
    P(R({\lambda})\le \alpha) \ge 1-\delta
    \label{eq:def_rcps}
    \end{equation}
%
%
%
Intuitively, in this work, this means the probability of observing the risk function $R \le \alpha$  is at least $1-\delta$ across repeated runs on different random data splits, when the set-valued function $\mathcal{T}(\lambda)$ is used to select which items can be sampled at a certain position. For example, in classification, $\mathcal{T}(\lambda)$ can be a set of classes with softmax probability higher than the threshold $\lambda$~\cite{angelopoulos2021gentle}.

To find proper thresholds for $\mathcal{T}(\lambda)$ to satisfy Eq.~\eqref{eq:rcps}, UCB based distribution-free risk control~\cite{bates2021distribution} needs the following assumptions:
%
    
    \noindent A1 \textit{Nesting Properties.} With larger threshold $\lambda$ (i) the prediction set $\mathcal{T}(\lambda)$ becomes smaller and (ii) the risk function is non-increasing:
    \begin{equation}
    \lambda < \lambda' \Rightarrow \mathcal{T}(\lambda) \subset \mathcal{T}(\lambda'), \Rightarrow R(\mathcal{T}(\lambda)) \ge R(\mathcal{T}(\lambda')).
        \label{eq:nesting}
    \end{equation}
    For example, in ranking, suppose we know the relevance, we can use $\mathcal{T}(\lambda)$ to filter out items with relevance lower than $\lambda$ for a position. In this case, as $\lambda$ increases, the set $\mathcal{T}(\lambda)$ becomes smaller, and the corresponding utility risk $R(\mathcal{T}(\lambda))$ is non-increasing. 
    
    \noindent A2 Existence of an upper confidence bound (UCB) $\hat{R}^+({\lambda})$ for the risk function, which satisfies
\begin{equation*}
P(R({\lambda}) \le \hat{R}^+({\lambda})) \ge 1 - \delta    
\end{equation*}


%

%

Under the aforementioned assumptions, distribution-free risk control~\citep{bates2021distribution} selects the threshold $\hat{\lambda}$ using the UCB $\hat{R}^+({\lambda})$ on a calibration dataset $\mathcal{Q}_{cal}$ s.t. $\mathcal{T}(\hat{\lambda})$ is a $(\alpha,\delta)$ risk-controlling prediction set. Intuitively, they select the $\lambda$ s.t. any $\lambda' \ge \lambda$ leads to UBC smaller than the desired level $\alpha$ as
\begin{equation}
    \hat{{\lambda}} = \inf \{ {\lambda} \in {\Lambda} : \hat{R}^+({\lambda'}) < \alpha, \forall {\lambda}' \ge  {\lambda} \},
    \label{eq:rcps}
\end{equation}

\noindent\textbf{P-value based Risk Control~\citep{angelopoulos2021learn}} extends distribution-free risk control to cases where the nesting properties are violated. It also allows multi-dimensional thresholds.
The crux of~\cite{angelopoulos2021learn} is hypothesis testing by the duality of p-values and concentration inequality, which selects a threshold by rejecting its corresponding null hypothesis $R(\bm{\lambda}) > \alpha$ with p-value smaller than $\delta$, where $\bm{\lambda}$ is the vector representing a multi-dimensional threshold. In Section~\ref{subsec:risk_control}, we will present concrete instantiation of both UBC and p-value based risk control for ranking.

It can be infeasible to control risk at every level of $\alpha$ for every data distribution~\citep{angelopoulos2021learn}.
For example, on real-world datasets with noisy relevance labels, guaranteeing $\text{NDCG@K} \ge 0.9$ may be unattainable given a pre-trained scoring function $f$, where risk control methods should abstain from returning a threshold.


%
%
%
%

\subsection{Generalized PL Ranking Model}
\label{subsec:TPL}

Here, we propose the Generalized PL (GPL) ranking model. With pretrained scoring functions from either deterministic or stochastic Learning-to-Rank (LTR) models, the GPL model facilitates an efficacious trade-off between utility and fairness at inference time\footnote{Nevertheless, it might be undesirable to use GPL with pre-trained scoring functions from stochastic LTR models due to the training cost in practice.}.
The GPL model builds connections between ranking models and set-valued functions, which enables distribution-free risk control for LTR.
With parameters of the set-valued functions obtained from risk control algorithms, the GPL model provides a guarantee on a specified risk function.



%
Suppose we have access to a risk control score $\tilde{s}(d)$ of each document $d$, which approximates the relevance of the item. In practice, we let $\tilde{s}(d)$ be a function of the predicted ranking score $s(d)$ from the pre-trained scoring function $f$. Specifically, we choose the probability to sample item $d$ at the first position in the PL model, $p(d|\emptyset,0)$.
More detailed discussion on the choice of risk control score can be found in Appendix~\ref{sec:rc_score}.
For each position $k$, the GPL ranking model uses a set-valued function $\mathcal{T}(\lambda_k)$ to select items whose predicted scores are high enough for position $k$, where $\lambda_k$ is the threshold parameter for position $k$:
\begin{equation}
\mathcal{T}(\lambda_k) = \{d|\tilde{s}_d \ge \lambda_k, \forall d \notin \by_{1:k-1} \},
\label{eq:set_valued_func}
\end{equation}
%
where $\by_{1:k-1}=\emptyset$ if $k=1$. 
For each position $k$, GPL creates a distribution of the items selected based on the set-valued function $\mathcal{T}(\lambda_k)$ defined in Eq.~\eqref{eq:set_valued_func} and then combines them to predict a distribution of rankings as:
\begin{equation}
\begin{split}
      p(d|\by_{1:k-1},\lambda_k) & = \frac{\mathds{1}(d\in \mathcal{T}(\lambda_k))\exp({s}_d/\tau)}{\sum_{d'\in \mathcal{T}(\lambda_k)} \exp({s}_{d'}/\tau)}, 
      \;\; \\ \pi(\by) & = \prod_{k=1}^K p(d|\by_{1:k-1},\lambda_k),
      \end{split}
      \label{eq:pseudo_dist_TPL}
\end{equation}

When $\lambda_k$ takes extreme values, the GPL model will reduce to the PL and the deterministic ranking model. Specifically, when $\lambda_k=0$ and $\lambda_k \ge \max(\{\Tilde{s}_d\}_{d\in \mathcal{D}^q\setminus \by_{1:k-1}} )$ for $k=1,...,K$, GPL is equivalent to PL and the deterministic ranking model, respectively.
We verify this empirically in Section~\ref{subsec:results} (see Fig.~\ref{fig:tradeoff_yahoo}). 
%



%
With thresholds, the GPL ranking model can adapt the prediction set size $|\mathcal{T}({\lambda_k})|$ for each position to achieve the goal described by the problem statement (Eq.~\eqref{eq:prob_state}).
 %
 For example, when there are two items with scores much higher than others, then GPL can select high thresholds for top-$2$ positions and use lower thresholds for other positions to guarantee utility while still achieving better amortized individual fairness than deterministic LTR models.
With the set-valued function $\mathcal{T}(\lambda_k)$, GPL can adapt the distribution of sampled rankings to achieve this goal at inference time. This is different from the PL model which cannot adjust the distribution of rankings at inference time.
Without thresholds, the PL ranking model is prone to sampling items with low relevance from $\mathcal{D}^q\setminus \by_{1:k-1}$ at the top positions when $k$ is small.
Deterministic LTR models cannot be adjusted at inference time either, which inherently predict a degenerate ranking distribution by design.
Next, we discuss the selection of thresholds by our distribution-free risk control framework in Section~\ref{subsec:risk_control}. 

%
%


%

%
%

%


%

\subsection{Risk Control for Generalized PL Model}
\label{subsec:risk_control}

Here, we describe our inference-time distribution-free risk control algorithm used to select thresholds for the GPL model to provide finite-sample guarantees on a given bounded risk function $R$.
Given a desired risk level $\alpha$ for a bounded risk function $R$ (e.g., NDCG@K),
our method uses the calibration set $\mathcal{Q}_{cal}$ to search for thresholds that lead to guaranteed utility.
%
%
Under the constraint that the risk $R$ is guaranteed, we aim to minimize another risk $R'$.
When $R=R_{util}$ and $R'=R_{fair}$ ($R=R_{fair}$ and $R'=R_{util}$), this is equivalent to maximizing (minimizing) the set size $|\mathcal{T}({\lambda_k})|$ by including items whose scores are at least $\lambda_k$ as candidates of position $k$.
Existing post-hoc methods for amortized individual fairness needs to solve additional optimization problems before inference (e.g., linear programming in~\cite{morik2020controlling}), which is less flexible than our method and cannot provide such guarantees.

\subsubsection{Threshold Selection via Distribution-free Risk Control.}
\label{subsubsec:threshold_selection}
We leverage distribution-free risk control~\citep{bates2021distribution,angelopoulos2021gentle} to select thresholds $\bm{\lambda}=[\lambda_1,...,\lambda_K]$ for the top-K positions s.t. the risk $R$ is highly likely to be not greater than a specified level $\alpha$, i.e., $P(R(\bm{\lambda}) \le \alpha) \ge 1-\delta$, where $R(\bm{\lambda})$ is short for $ R(\pi(\bm{\lambda}))$. 
%
%

The distribution-free risk control algorithm that selects thresholds for GPL works as follows.
First, we specify a search space $\Lambda$ for the thresholds. Each value $\bm{\lambda}\in \Lambda$ corresponds to a null hypothesis
$R(\bm{\lambda}) > \alpha$.
Second, for each value of $\bm{\lambda}$, we perform p-value or UCB based null hypothesis testing on the calibration set $\mathcal{Q}_{cal}$ which is assumed to share the same distribution with the test set $\mathcal{Q}_{test}$~\citep{angelopoulos2021learn}.
For both testing approaches, we start with computing the risk of $R(\bm{\lambda})$ for each of the thresholds $\bm{\lambda}$ in $\Lambda$ by simply performing inference on the calibration set.

\noindent\textbf{P-value based Testing.} We can also obtain $\hat{\Lambda}$ by p-value based hypothesis testing as
\begin{equation}
    \hat{\Lambda} = \{\bm{\lambda} \in \Lambda | p(\bm{\lambda},\alpha) < \delta\},
    \label{eq:p-value}
\end{equation}
where the p-value $p(\bm{\lambda},\alpha)$ can be derived from concentration inequalities such as Hoeffding~\cite{hoeffding1994probability} and Bentkus~\cite{bentkus2004hoeffding}.

\noindent\textbf{UCB based Testing.} Given an upper confidence bound (UCB) for the risk $R(\bm{\lambda})$ of each $\bm{\lambda}$, denoted by $\hat{R}^+(\bm{\lambda},\delta)$ (see details in Section~\ref{subsubsec:rc_ranking}), our method obtains $\hat{\Lambda}$, a set of $\bm{\lambda}$ which leads to rejection of the null hypothesis by UCB based hypothesis testing as
\begin{equation}
\hat{\Lambda} = \{\bm{\lambda} \in \Lambda | \hat{R}^+(\bm{\lambda},\delta) < \alpha\}    
\end{equation}

Finally, we choose the threshold from $\hat{\Lambda}$ to minimize $R'$ as
\begin{equation}
    \hat{\bm{\lambda}} = {\arg\min}_{{\bm{\lambda}\in \hat{\Lambda}}}\; R'(\bm{\lambda})
    \label{eq:optimize_thres}
\end{equation}

However, the computation can be inhibitive if brute-force grid search is performed to test all possible values of $\bm{\lambda}$ from a predefined grid with $M$ values for each $\lambda_k$. This requires computing $R_{util}(\bm{\lambda}$ on the calibration set and the UCB $\hat{R}^+(\bm{\lambda},\delta)$ for  $M^K$ times.
In this work, we overcome this issue by limiting the search space of $\bm{\lambda}$.
%
Intuitively, the threshold of a lower position (e.g., $k+1$) should not be greater than that of a higher position (e.g., $k$), i.e., $\lambda_{k} \ge \lambda_{k+1}$, because the impact of ranking an item with low relevance at position $k+1$ is smaller than that at position $k$ with NDCG@K as the metric.
We simply use a scaling factor $\zeta  \in (0,1]$ to control the position-wise decrease as $\lambda_{k+1} = \zeta \lambda_{k}$.
Thus, with this limited search space, we can solve Eq.~\eqref{eq:optimize_thres} by simply selecting the $\bm{\lambda} \in \hat{\Lambda}$ with $\lambda_1$ leading to minimal risk $R'(\bm{\lambda})$.
This limited search space allows highly efficient distribution-free risk control to select thresolds for the GPL ranking model.
Moreover, this design with a single threshold parameter $\lambda_1$ also makes it possible to satisfy the assumed nesting property for the UCB based testing.
In Section~\ref{subsubsec:scaling_factor}, we show empirical analysis on the scaling factor $\zeta$.

\subsubsection{Distribution-free Risk Control for Ranking.}
\label{subsubsec:rc_ranking}
Here, we present concrete instantiations of the UCB $\hat{R}^+(\bm{\lambda},\delta)$ based and p-valued based distribution-free risk control~\cite{angelopoulos2021learn,bates2021distribution} for ranking, which returns a set of risk-controlling thresholds $\hat{\Lambda}$.

\noindent\textbf{P-value based Testing.}
We adopt the widely adopted Hoeffding-Benktus (HB) inequality~\citep{bates2021distribution,bentkus2004hoeffding,hoeffding1994probability} for risk functions bounded in $[0,1]$ based on the duality of concentration inequality and p-values in hypothesis testing. HB inequality combines the two inequalities by taking the minimum of the p-values from them. The p-value associated with HB inequality is a function of risk ${R}(\lambda)$ computed on the calibration set and the number of queries in the calibration set $|\mathcal{Q}_{cal}|$ as:
\begin{equation}
\begin{split}
    p^{HB}(\bm{\lambda},\alpha) = \min (\exp(-|\mathcal{Q}_{cal}|h_1(\min(R(\bm{\lambda}),\alpha),\alpha)), \\ \exp{(1)} \times p(Bin(n,\alpha)\le \lceil |\mathcal{Q}_{cal}|R(\bm{\lambda})\rceil)),   
\end{split}
    \label{eq:p_value_HB}
\end{equation}
where $h_1(a,b) = a \log (\frac{a}{b})+(1-a)\log(\frac{1-a}{1-b})$, $Bin(n,\alpha)$ is the Binomial distribution and $\lceil a \rceil$ takes the ceiling of the scalar $a$.
Given the Hoeffding Benktus p-values computed by Eq.~\eqref{eq:p_value_HB}, we can obtain the set of selected thresholds $\hat{\Lambda}=\{\bm{\lambda}\in \Lambda| p^{HB}(\bm{\lambda}) < \delta \}$. Then, we take the $\bm{\lambda} \in \hat{\Lambda}$ with minimal $\lambda_1$ as it heuristically minimizes the other risk $R'$.

Second, with the UCB based testing, we adopt a theory-backed UCB based on the Dvoretzky–Kiefer–Wolfowitz-Massart (DKWM) inequality~\citep{massart1990tight,dvoretzky1956asymptotic} which works for discrete risk functions that only take a finite number of values (e.g., $R=R_{util}=1-\text{NDCG@K}$). Note that one might also adopt the UCBs introduced in~\citep{bates2021distribution} if the risk function is satisfies corresponding assumptions.   
\begin{equation}
\hat{R}^+(\bm{\lambda},\delta) = R({\bm{\lambda}})+ \texttt{const.} \cdot \sqrt{\frac{\ln (2/\delta)}{2\cdot |\mathcal{Q}_{cal}|}},
\label{eq:DKWM_UCB}
\end{equation}
where $\texttt{const.} $ is a constant that depends on the set of loss values -- we specify the details and provide the proof in Appendix~\ref{sec:proof}.
As mentioned before, with such a UCB, we can select the threshold $\hat{\bm{\lambda}}$ by using Eq.~\eqref{eq:p-value} and Eq.~\eqref{eq:optimize_thres}. Thus, our method can provide the guarantee by Theorem 1 of~\citep{bates2021distribution}. 

For UCB based testing, when $R(\bm{\lambda})$ does not satisfy the nesting assumption (Eq.~\eqref{eq:nesting}), we can modify the risk function to make the assumption satisfied.
By the aforementioned limited search space of thresholds, we can simply create a new risk function as $\tilde{R}(\bm{\lambda})= \max_{t \le  \lambda_1} (\bm{\lambda})$ which upperbounds the original utility risk function and satisfies the nesting property. Intuitively, when $R(\bm{\lambda}) < R(\bm{\lambda}')$ with $\lambda_1 > \lambda_1'$, it violates the nesting property. $\tilde{R}(\bm{\lambda})$ can avoid this violation by its definition.
Then, we can apply the UCBs on the newly obtained utility risk function $\tilde{R}(\bm{\lambda})$.

We summarize our the calibration phase of the proposed method ISRR in Algorithm~\ref{algo:ISRR}.

\begin{algorithm}
\caption{Calibration Phase of ISRR}
\label{algo:ISRR}
\begin{algorithmic}[1]
\Require $f$: a scoring function from a pre-trained Learning-to-Rank (LTR) model, 
        $\mathcal{Q}_{cal}$: calibration set, $\Lambda$: parameter space, 
        $\alpha$: risk level, $\delta$: coverage, 
        $R$: risk function, $R'$: secondary risk function
\Ensure $\hat{\boldsymbol{\lambda}}$ that guarantees risk $P(R(\hat{\boldsymbol{\lambda}}) < \alpha) > 1 - \delta$ 
        and minimizes risk $R'(\boldsymbol{\lambda})$, or abstains if $\hat{\Lambda}=\emptyset$

\Procedure{Calibration}{$\Lambda$, $\mathcal{Q}_{cal}$}
    \For{each $\boldsymbol{\lambda} \in \Lambda$}
        \For{each query $q \in \mathcal{Q}_{cal}$}
            \State sample $L$ rankings from GPL model with $\boldsymbol{\lambda}$ and evaluate them
        \EndFor
        \State $R(\boldsymbol{\lambda}) \gets$ compute mean risk of GPL with $\bm{\lambda}$ on $\mathcal{Q}_{cal}$
    \EndFor
    
    \State $\hat{\Lambda} \gets$ find a subset of $\Lambda$ with guaranteed risk using p-value (Eq. (8)) or UCB (Eq. (9))
    \If{$|\hat{\Lambda}| == 0$}
        \State \Return None
    \Else
        \State $\hat{\boldsymbol{\lambda}} \gets$ $\arg\min_{\boldsymbol{\lambda} \in \hat{\Lambda}} R'(\boldsymbol{\lambda})$ (Eq. (10))
        \State \Return $\hat{\boldsymbol{\lambda}}$
    \EndIf
\EndProcedure
\end{algorithmic}
\end{algorithm}



%
%



\section{Experiments}

In this section, we perform experiments on three popular LTR benchmark datasets with pre-trained scoring functions from both deterministic and stochastic LTR models.
Specifically, we consider a practical situation where we start from the score functions of deterministic or stochastic LTR models which are pre-trained to optimize utility. Then, our task is to improve fairness by minimizing $R'=R_{fair}$ with utility risk $R=R_{util}$ guaranteed to be not greater than a specified level with high probability.

We answer the following research questions:
 \textbf{RQ1}: Can our proposed GPL achieve trade-off between utility and amortized indivudal fairness at inference time? \textbf{RQ2}: Does GPL generalize the PL and deterministic ranking models? \textbf{RQ3}: Can \ourmodel achieve high coverage on the risk function and improve amortized indivudal fairness at the same time? \textbf{RQ4}: Is ISSR much more efficient than existing stochastic LTR models? \textbf{RQ5}: What is the impact of the scaling factor $\zeta$ in ISSR on the evaluation metrics?


\subsection{Experimental Setup}

\noindent\textbf{Datasets.}
We consider two popular publicly available datasets for LTR: Yahoo!Webscope (Yahoo)~\citep{chapelle2011yahoo}, MSLR-WEB30k (MSLR)~\citep{qin2013introducing} and Istella-S~\citep{dato2016fast}. 
%
\begin{table}[tbh!]
\caption{Statistics of the benchmark datasets}
\label{tab:data}
\scriptsize
\begin{tabular}{c|ccc|c}
\hline
            & \multicolumn{3}{c|}{\# queries (\# items)}                                                          & \# features \\ \hline\hline
            & \multicolumn{1}{c|}{training}          & \multicolumn{1}{c|}{validation}     & original test    &           \\ \hline
Yahoo       & \multicolumn{1}{c|}{19,944 (473,134)}  & \multicolumn{1}{c|}{6,983 (165,660)} & 2,994 (71,083)   & 700       \\ \hline
MSLR-WEB30K & \multicolumn{1}{c|}{18,919 (2,270,296)} & \multicolumn{1}{c|}{6,306 (753,611)} & 12,581 (747,218)  & 136       \\ \hline
Istella-S   & \multicolumn{1}{c|}{19,245 (2,043,304)} & \multicolumn{1}{c|}{7,211 (681,250)} & 6,528 (684,076) & 220       \\ \hline
\end{tabular}
\vspace{-10pt}
\end{table}
Table~\ref{tab:data} shows the statistics describing the widely used datasets for evaluating LTR models. We observe that Yahoo has more features and MSLR has more queries and much more items per query.
These datasets consist of queries, their associated documents, and relevance labels in $0-4$ indicating the expert-judged relevance between an item and a query. Each feature vector represents a query-document pair.

Similar to~\citep{angelopoulos2022recommendation}, to compute the coverage rate, we repeat the experiment $50$ times by randomly splitting the original test set into calibration (25\%) and test sets (75\%).
The scoring functions are pre-trained on the training set and model selection is done by maximizing NDCG@5 on the validation set.
Then, for risk control, the threshold ${\lambda}$ is selected based on UCB or p-value computed on the calibration set.
We compare the proposed \ourmodel with the deterministic and PL ranking model with mean and standard deviation of the test performance over these 50 runs.

The NDCG@K reported in this work may look lower than those in the literature because we follow~\citep{bruch2020stochastic} to ignore all queries with no relevant documents for fair evaluation.
%
%
It avoids arbitrarily assigning $\text{NDCG@K}=1$ or $0$ to such queries with any predicted ranking, to prevent unfair comparisons.

\noindent\textbf{Scoring Functions.}
We use \cat~\citep{prokhorenkova2018catboost}, \lgb \ (LGB)~\citep{ke2017lightgbm} and Neural Network (NN) as the pre-trained scoring function.
The NN is a three-layer MLP with sigmoid activation trained with LambdaLoss~\citep{wang2018lambdaloss}. 
On top of the pre-trained scoring functions, we apply the ranking models (deterministic, PL and GPL). More details about the experimental setup can be found in Appendix~\ref{sec:setup_details}.
To make the coverage results more comprehensive, we also apply our method to the state-of-the-art stochastic LTR models that train scoring function on the top of the PL model, including PL-Rank-3~\cite{oosterhuis2022learning}, StochasticRank~\citep{ustimenko2020stochasticrank}, and Policy Gradient~\cite{singh2019policy}, the results for which is in Appendix~\ref{sec:detailed_results}.

%

%

%
%
%

%

%

\subsection{Experimental Results}
\label{subsec:results}



\subsubsection{GPL's Trade-off between Utility and Fairness (RQ1-2).}
\begin{figure*}[tbh!]
  \centering
  \begin{subfigure}{0.33\textwidth}
    \centering
    \includegraphics[width=\linewidth]{figs/yahoo/interpolation_model_name_cat-query-rmse_scaling_1.0ndcg_sq_fair}
    \caption{CatBoost on Yahoo}
\label{fig:tradeoff_yahoo_cat}
  \end{subfigure}%
  \hfill
    \begin{subfigure}{0.33\textwidth}
  \includegraphics[width=\linewidth]{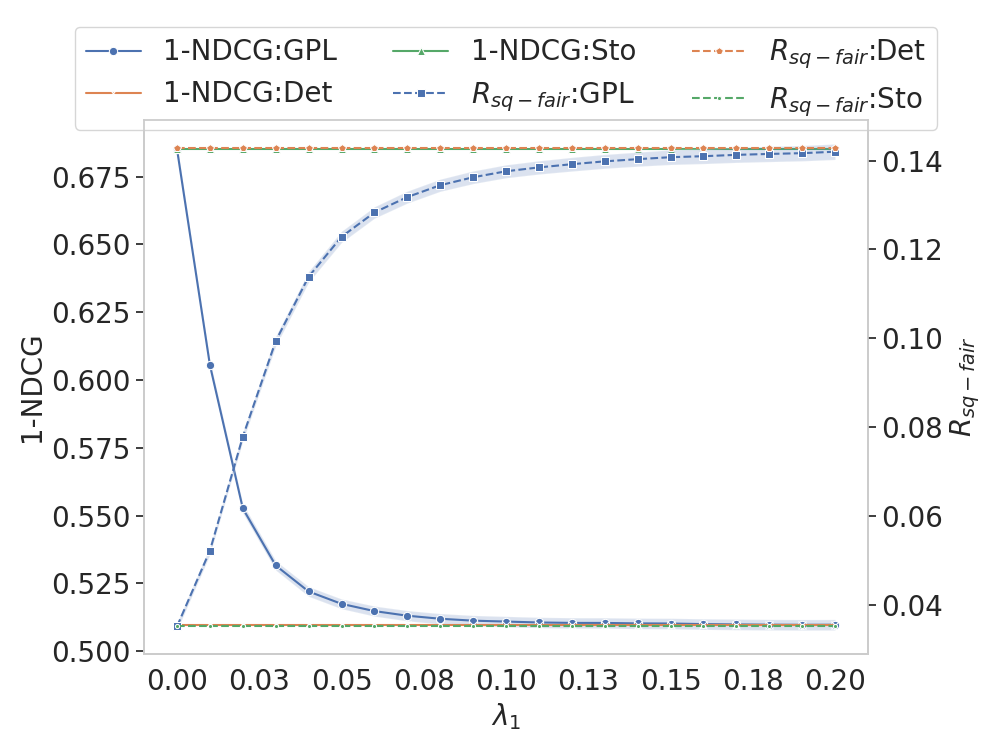}
    \caption{CatBoost on MSLR-WEB30K}
    \label{fig:tradeoff_mslr_cat}
  \end{subfigure}%
  \hfill
   \begin{subfigure}{0.33\textwidth}
  \includegraphics[width=\linewidth]{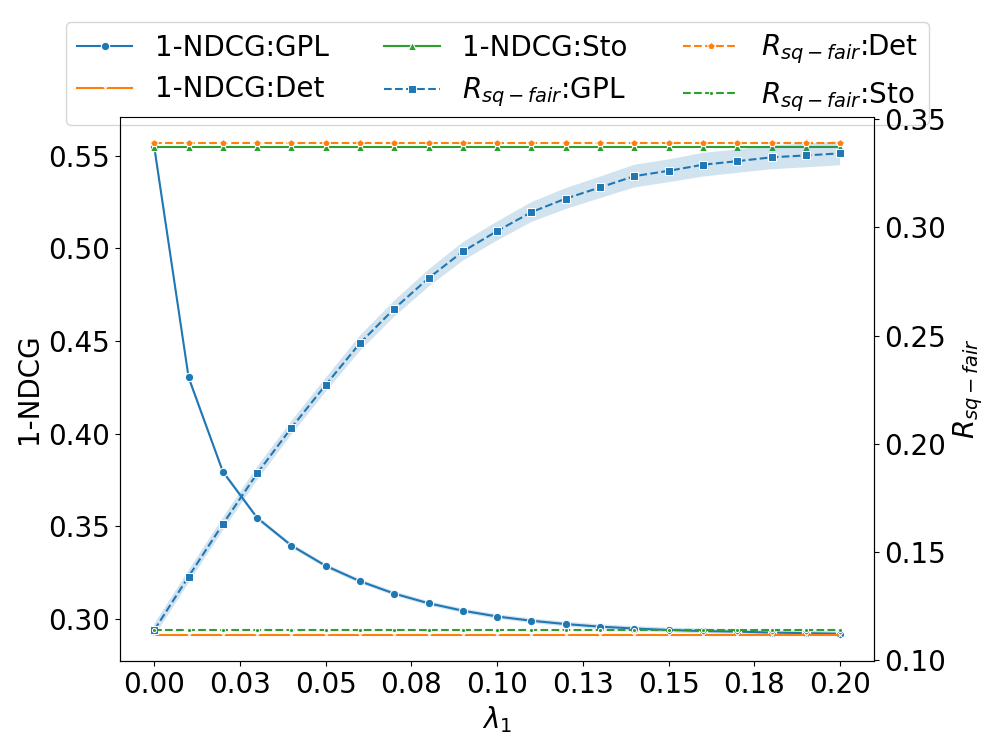}
    \caption{CatBoost on Istella-S}
\label{fig:tradeoff_istella_cat}
  \end{subfigure}%
  \hfill
  \begin{subfigure}{0.33\textwidth}
    \centering
\includegraphics[width=\linewidth]{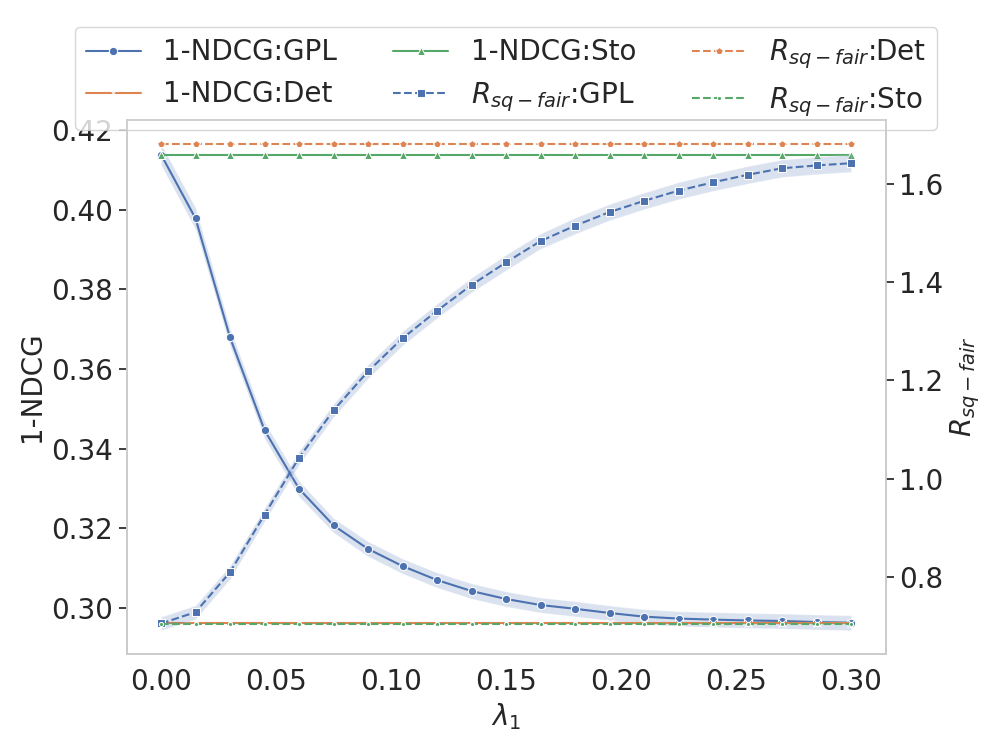}
    \caption{NN on Yahoo}%
  \end{subfigure}%
  \hfill
  \begin{subfigure}{0.33\textwidth}
    \centering
\includegraphics[width=\linewidth]{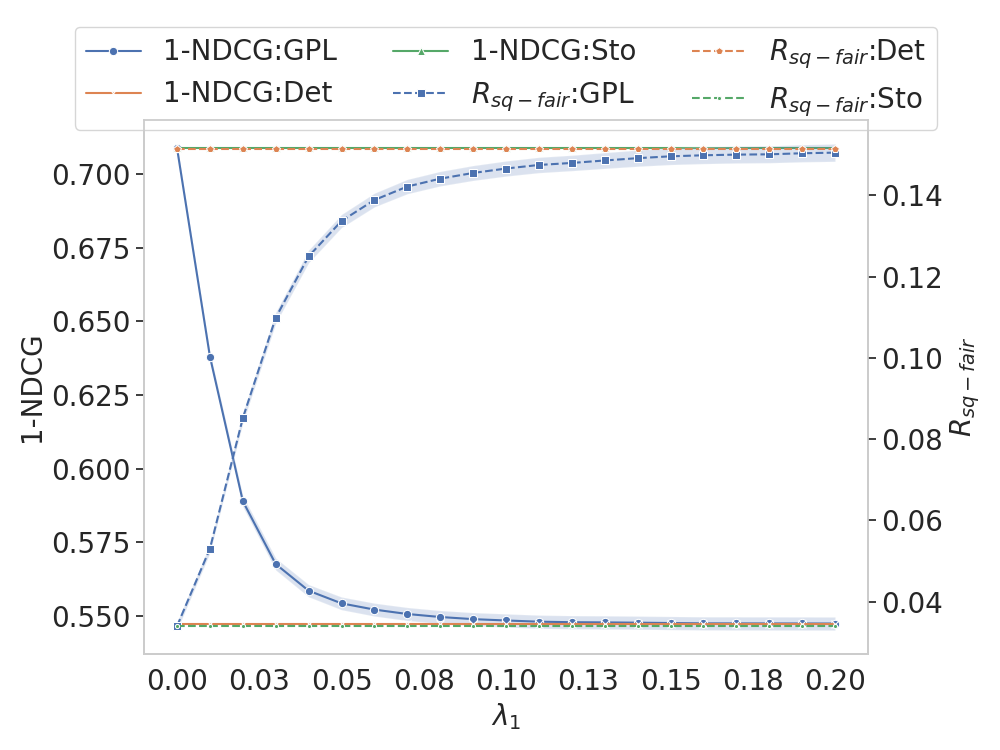}
    \caption{NN on MSLR-WEB30K}
\label{fig:tradeoff_mslr_NN}
  \end{subfigure}%
   \begin{subfigure}{0.33\textwidth}
    \centering
\includegraphics[width=\linewidth]{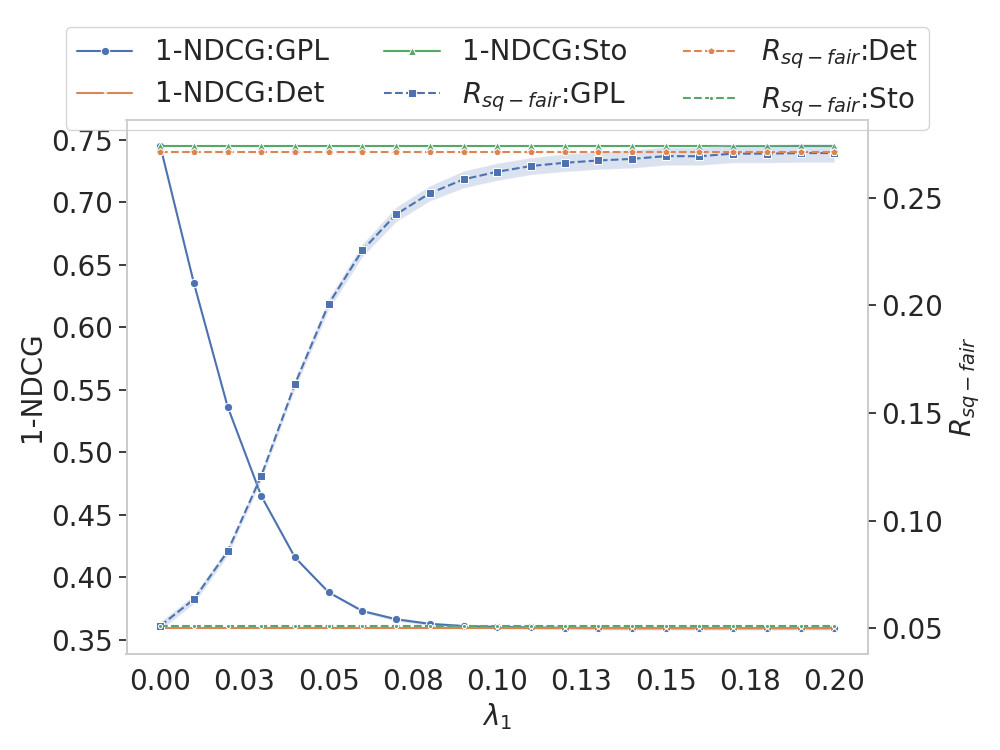}
    \caption{NN on Istella-S}
\label{fig:tradeoff_mslr_NN}
  \end{subfigure}%
  \caption{Trade-off between utility and fairness of the proposed GPL ranking model.
    The x-axis is the threshold $\lambda$, the left (right) y-axis is the risk $R_{{util}}=1-\text{NDCG@5}$ (the disparity $R_{sq-fair}$). Det and Sto are the deterministic and the PL ranking models. The shade is one standard deviation. As $\lambda$ increases, GPL only includes items with higher scores, leading to lower risk and higher disparity. Results with other models are in Fig.~\ref{fig:tradeoff_full} of Appendix~\ref{sec:detailed_results}.}
  \label{fig:tradeoff_yahoo}
\end{figure*}

Here, we first verify that the proposed GPL ranking model achieves effective utility fairness trade-off and reduces to the PL and deterministic models with certain values of $\lambda$ (\textbf{RQ1}). Specifically, we show how GPL's utility and fairness varies across different values of threshold $\lambda$. Note that we did not apply risk control in this experiment.
For evaluation, we adopt the widely used NDCG@5 as the utility metric.
For amortized indivudal fairness, we follow~\citep{oosterhuis2021computationally} to instantiate Eq.~\eqref{eq:def_fair} with $\ell(a,b)=(a-b)^2$ to obtain mean squared disparity $R_{sq-fair}$.
It measures how the exposure of an item is different from being proportional to its relevance.
As shown in Fig.~\ref{fig:tradeoff_yahoo},
when the thresholds increases, the utility (risk) of the GPL ranking model increases (decreases) while the disparity measure increases. 
In addition, results verify the claim that the GPL model can reduce to the PL and deterministic model (\textbf{RQ2}).
When $\lambda_k=0,k=1,...,K$, the GPL model reduces to the PL ranking model (Sto). When $\lambda \ge \max(\Tilde{s}_d)$ (e.g., $\lambda_1 \ge 0.4$ on Yahoo for CatBoost in Fig.~\ref{fig:tradeoff_yahoo_cat}), GPL reduces to the deterministic ranking model (Det).

\begin{table}[htbp]
\scriptsize
\caption{Coverage and FairGain results with thresholds $\lambda$ selected by the p-value of Hoeffding-Benktus. All pre-trained scoring functions are optimized for utlity. Similar results can be found in Table~\ref{tab:complete_coverage_results} in Appendix~\ref{sec:detailed_results} for pre-trained scoring functions from stochastic LTR models optimized for utlity.}
\label{tab:coverage}
  \setlength{\tabcolsep}{4pt} %
\begin{tabular}{|c|ccccc|}
\hline
         & \multicolumn{1}{c|}{Coverage} & \multicolumn{1}{c|}{1-$\alpha$} & \multicolumn{1}{c|}{NDCG@5} & \multicolumn{1}{c|}{FairGain} & \# Abstention \\ \hline
         & \multicolumn{5}{c|}{Yahoo}                                                                                                                                                  \\ \hline
CatBoost+ISRR & \multicolumn{1}{c|}{100.00\%} & \multicolumn{1}{c|}{0.687}      & \multicolumn{1}{c|}{0.727}  & \multicolumn{1}{c|}{23.99\%}                               & 2              \\ \hline
LightGBM+ISRR & \multicolumn{1}{c|}{100.00\%} & \multicolumn{1}{c|}{0.687}      & \multicolumn{1}{c|}{0.727}  & \multicolumn{1}{c|}{20.77\%}                               & 0              \\ \hline
NN+ISRR       & \multicolumn{1}{c|}{100.00\%} & \multicolumn{1}{c|}{0.641}      & \multicolumn{1}{c|}{0.673}  & \multicolumn{1}{c|}{35.05\%}                               & 0              \\ \hline
         & \multicolumn{5}{c|}{MSLR}                                                                                                                                                   \\ \hline
CatBoost+ISRR & \multicolumn{1}{c|}{100.00\%} & \multicolumn{1}{c|}{0.449}      & \multicolumn{1}{c|}{0.481}  & \multicolumn{1}{c|}{16.08\%}                               & 0              \\ \hline
LightGBM+ISRR & \multicolumn{1}{c|}{100.00\%} & \multicolumn{1}{c|}{0.449}      & \multicolumn{1}{c|}{0.480}  & \multicolumn{1}{c|}{13.29\%}                               & 2              \\ \hline
NN+ISRR      & \multicolumn{1}{c|}{100.00\%} & \multicolumn{1}{c|}{0.405}      & \multicolumn{1}{c|}{0.430}  & \multicolumn{1}{c|}{26.14\%}                               & 0              \\ \hline
         & \multicolumn{5}{c|}{Istella-S}                                                                                                                                              \\ \hline
CatBoost+ISRR & \multicolumn{1}{c|}{100.00\%} & \multicolumn{1}{c|}{0.637}      & \multicolumn{1}{c|}{0.667}  & \multicolumn{1}{c|}{35.29\%}                               & 0              \\ \hline
LightGBM+ISRR & \multicolumn{1}{c|}{100.00\%} & \multicolumn{1}{c|}{0.637}      & \multicolumn{1}{c|}{0.667}  & \multicolumn{1}{c|}{36.15\%}                               & 0              \\ \hline
NN+ISRR      & \multicolumn{1}{c|}{100.00\%} & \multicolumn{1}{c|}{0.572}      & \multicolumn{1}{c|}{0.609}  & \multicolumn{1}{c|}{27.31\%}                               & 0              \\ \hline
\end{tabular}
\vspace{-5pt}
\end{table}
\subsubsection{Utility Guarantee and Fairness Improvement (RQ3).}
For evaluation of utility guarantee in terms of coverage and fairness improvement, we let $U^*$ and $R_{sq-fair}^*$ be the NDCG@5 and mean squared disparity of a given pre-trained LTR model optimized for utility.

To evaluate the coverage, we repeat the experiment on $50$ random splits of test and calibration sets from the original test set. We report the \textbf{coverage rate} $ \sum_{t=1}^T \mathds{1}(R_{util}(\hat{\lambda})\le\alpha) /T$ computed on the test sets with thresholds $\hat{\lambda}$ which are selected by risk control.
We define \textbf{fairness gain (FairGain)} as $1-\frac{R_{sq-fair}}{R^*_{sq-fair}}$. It equals to $1$ ($0$) if our method leads to perfectly fair rankings (no improvement).

We choose $1-\alpha=0.9U^*$ based on the performance of the original LTR model.
With results shown in Table~\ref{tab:coverage}, we make the following observations:
    With thresholds selected based on the p-values of Hoeffding-Benktus, in at least 48 out of 50 runs ($\le 2$ abstentions), our method achieves 100\% coverage on the utility guarantee.
    At the same time, our method improves fairness significantly compared to the determinstic ranking models with at least 20.77\%, 13.29\%, and 27.31\% improvement in terms of the mean squared disparity $R_{sq-fair}$.
    Note that 100\% coverage does not mean it is perfect when our target is 95\% coverage with $\delta=0.05$.
    As in the utility fairness trade-off, once the coverage reaches the specified level, our main focus is to optimize fairness.
    Moreover, when the method is too conservative, it will result in coverage rate much higher than the desired level, and also higher chance to abstain from selecting any threshold.
    In practice, when the risk control method abstains from selecting any thresholds, we can set the thresholds $\lambda_k=1, k=1,...,K$ to make GPL reduce to the deterministic ranking model.
Coverage and FairGain results for UCB based testing with DKWM inequality is in Appendix~\ref{sec:detailed_results}.

%

%


\begin{figure}[tbh!]
  \centering
  
  \begin{subfigure}{0.23\textwidth}
    \centering
    \includegraphics[width=\linewidth]{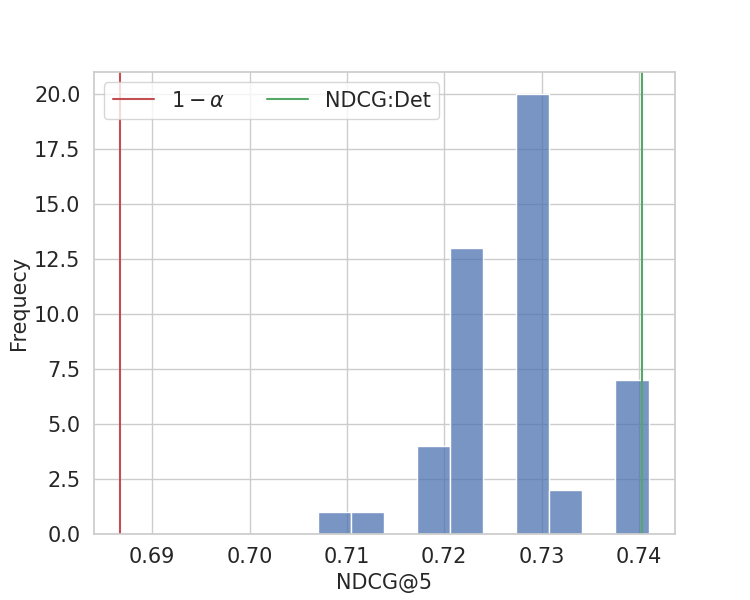}
\subcaption{\cat \ on Yahoo}
    \label{fig:coverage_cat_yahoo}
  \end{subfigure}%
  \hfill
  \begin{subfigure}{0.23\textwidth}
    \centering
    \includegraphics[width=\linewidth]{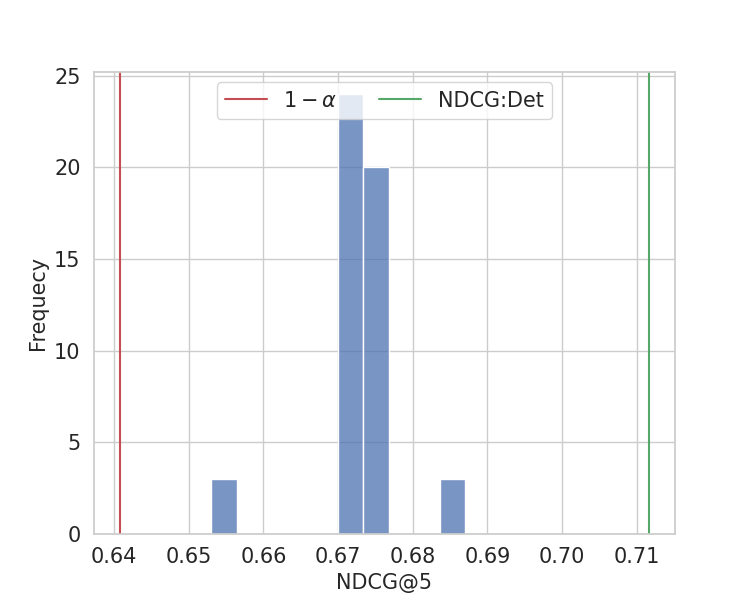}
    \caption{NN on Yahoo}
\label{fig:coverage_NN_yahoo}
  \end{subfigure}%
  \caption{Distribution of NDCG@5 achieved by \ourmodel over $50$ runs with $\lambda$ selected by Hoeffding-Benktus. 
  Results for other datasets, and \lgb \ can be found in Appendix~\ref{sec:detailed_results}.
  The red (green) vertical line is the desired NDCG@5 level $1-\alpha = 0.9 U^*$. $U^*$ is the NDCG@5 of the pre-trained LTR model.}
  \label{fig:coverage_yahoo_cat_nn}
\end{figure}

In addition, Fig.~\ref{fig:coverage_yahoo_cat_nn} show two examples of the actual distribution of the utility (NDCG@5) of our proposed method over the 50 runs on the Yahoo dataset. We can clearly observe that our method leads to 100\% coverage. However, this also implies that HB is still conservative as all the obtained NDCG@5 are much higher than the specified level $1-\alpha$.
Fig.~\ref{fig:coverage} in Appendix~\ref{sec:detailed_results} shows the complete results for distribution of $\text{NDCG@5}$ with thresholds selected by Hoeffding-Benktus and the DKWM iequality.
 %



\subsubsection{Running Time Results (RQ4).}
\label{subsubsec:running_time}
\begin{table*}[htbp]
\caption{Average running time in seconds for our method (LambdaLoss+ISRR) and stochastic LTR models. We report the training time for $100$ epochs. Stochastic LTR models use $100$ sampled rankings per query and risk control takes $21$ sets of thresholds as in other experiments.}
\label{tab:run_time}
\scriptsize
\begin{tabular}{|c|ccc|ccc|ccc|}
\hline
                        & \multicolumn{3}{c|}{Yahoo}                                                   & \multicolumn{3}{c|}{MSLR}                                                    & \multicolumn{3}{c|}{Istella-S}                                               \\ \hline
                        & \multicolumn{1}{c|}{Training} & \multicolumn{1}{c|}{Risk Control} & Total    & \multicolumn{1}{c|}{Training} & \multicolumn{1}{c|}{Risk Control} & Total    & \multicolumn{1}{c|}{Training} & \multicolumn{1}{c|}{Risk Control} & Total    \\ \hline
LambdaLoss+ISRR & \multicolumn{1}{c|}{222.947}  & \multicolumn{1}{c|}{95.023}       & 317.970  & \multicolumn{1}{c|}{1806.129} & \multicolumn{1}{c|}{141.017}      & 1947.146 & \multicolumn{1}{c|}{599.524}  & \multicolumn{1}{c|}{135.081}      & 734.605  \\ \hline
PL-Rank-3               & \multicolumn{1}{c|}{2876.254} & \multicolumn{1}{c|}{--}           & 2876.254 & \multicolumn{1}{c|}{6101.400} & \multicolumn{1}{c|}{--}           & 6101.400 & \multicolumn{1}{c|}{1448.598} & \multicolumn{1}{c|}{--}           & 1448.598 \\ \hline
Policy Gradient         & \multicolumn{1}{c|}{2999.316} & \multicolumn{1}{c|}{--}           & 2999.316 & \multicolumn{1}{c|}{5172.069} & \multicolumn{1}{c|}{--}           & 5172.069 & \multicolumn{1}{c|}{2498.054} & \multicolumn{1}{c|}{--}           & 2498.054 \\ \hline
StochasticRank          & \multicolumn{1}{c|}{3700.335} & \multicolumn{1}{c|}{--}           & 3700.335 & \multicolumn{1}{c|}{6291.146} & \multicolumn{1}{c|}{--}           & 6291.146 & \multicolumn{1}{c|}{2719.025} & \multicolumn{1}{c|}{--}           & 2719.025 \\ \hline
\end{tabular}
\end{table*}

Here, we compare the running time of \ourmodel with the state-of-the-art stochastic LTR models including PL-Rank-3~\cite{oosterhuis2022learning}, PolicyGradient~\cite{singh2019policy} and StochasticRank~\cite{ustimenko2020stochasticrank}.
For fair comparision, we perform all the experiments on the same type of nodes of an internal CPU cluster with Intel Xeon Platinum 8260 @ 2.40GHz CPUs (24 cores) and 30 GB RAM, using the same NN based scoring function for all methods. For ISRR, we report the running time of (1) training a deterministic LTR model using the deterministic implementation of LambdaLoss~\cite{wang2018lambdaloss} for $100$ epochs, (2) risk control including inference on calibration set for $21$ sets of thresholds and threshold selection.
For stochastic LTR models, we report the running time of training $100$ epochs.

As shown in Table~\ref{tab:run_time}, we can observe that, across all three datasets, our method, including training a determinstic LTR model for $100$ epochs and performing risk control, is much more efficient than training any of the stochastic LTR models for $100$ epochs.

Note that for training on each query, some stochastic LTR models (e.g., PL-Rank-3~\cite{oosterhuis2022learning}) may have better time complexity for a single sampled ranking list, however, they need to sample $100$ to $1,000$ ranking lists for each query to accurately approximate the gradient as in~\cite{oosterhuis2022learning}.
While determinisitc LTR models (e.g., LambdaLoss~\cite{wang2018lambdaloss}) only need to compute the gradient on a single ranking list.

%
%

\subsubsection{Impact of Scaling Factor (RQ5).}
\label{subsubsec:scaling_factor}
The scaling factor $\zeta$ is important as it controls how fast the threshold decreases per position.
Here, we study the impact of the scaling factor on the performance of ISRR.
The scaling factor $\zeta\in (0,1]$ controls the degree of decrease per position of the threshold as $\lambda_{k+1} = \zeta \lambda_k $.
In particular, we select $\zeta \in [0.1,0.3,0.5,0.7,0.9,1.0]$ and report results of the utility fairness trade-off, coverage and fairness improvement.
With limited space, we report results for CatBoost on Yahoo dataset.

\begin{figure*}[htbp]
  \centering
  \begin{subfigure}{0.33\textwidth}
    \centering
    \includegraphics[width=\linewidth]{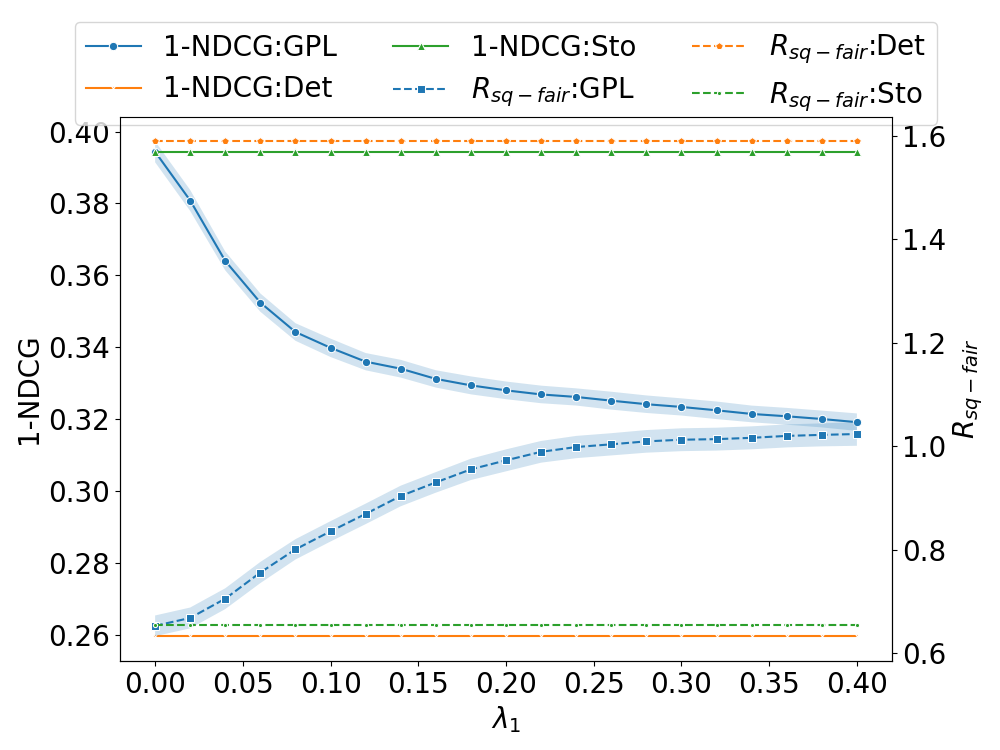}
    \caption{$\zeta=0.1$ with CatBoost on Yahoo.}
  \end{subfigure}%
  \hfill
  \begin{subfigure}{0.33\textwidth}
    \centering
\includegraphics[width=\linewidth]{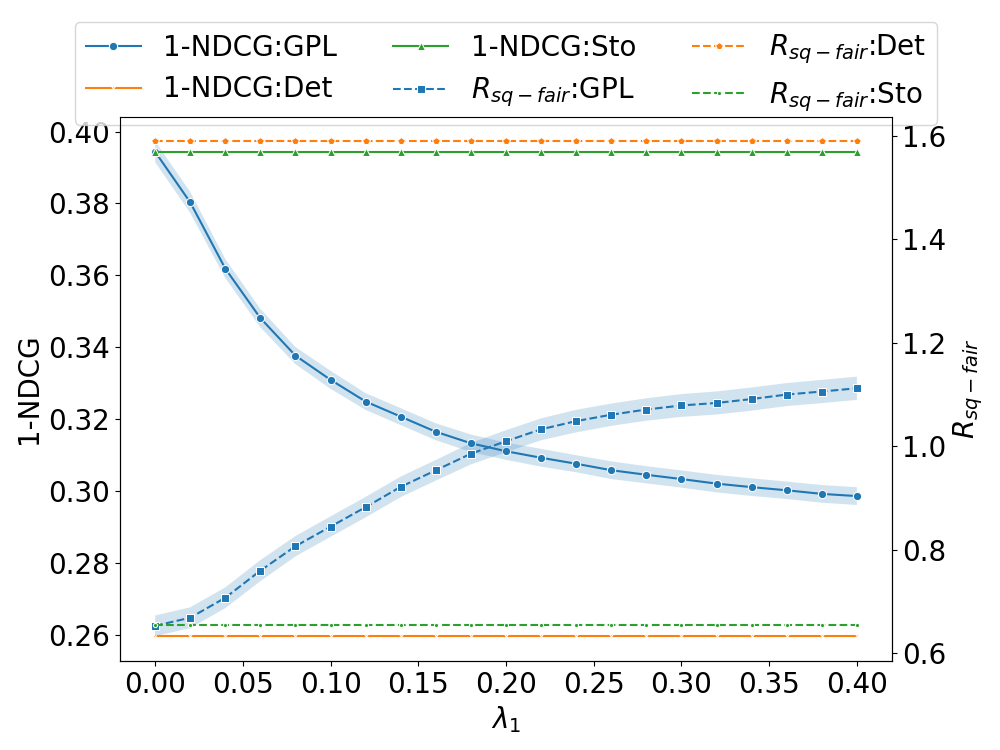}
    \caption{$\zeta=0.3$ with CatBoost on Yahoo.}%
  \end{subfigure}%
  \hfill
  \begin{subfigure}{0.33\textwidth}
  \includegraphics[width=\linewidth]{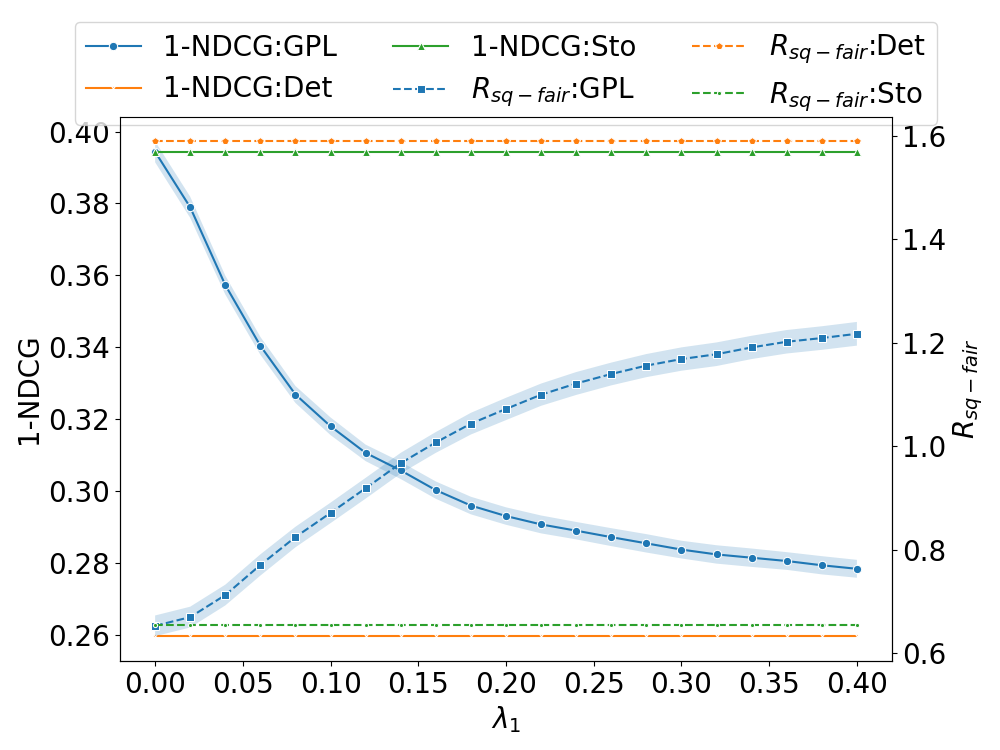}
    \caption{$\zeta=0.5$ with CatBoost on Yahoo.}
    \label{fig:tradeoff_mslr_cat}
  \end{subfigure}%
  \hfill
  \begin{subfigure}{0.33\textwidth}
    \centering
\includegraphics[width=\linewidth]{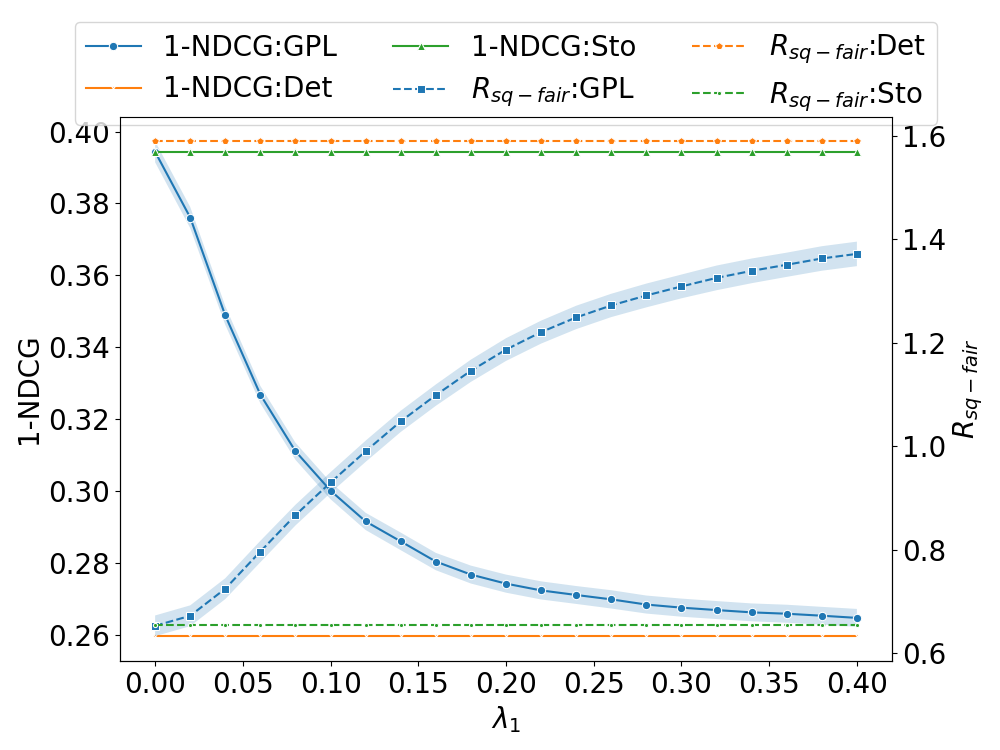}
    \caption{$\zeta=0.7$ with CatBoost on Yahoo.}
  \end{subfigure}%
    \begin{subfigure}{0.33\textwidth}
    \centering
\includegraphics[width=\linewidth]{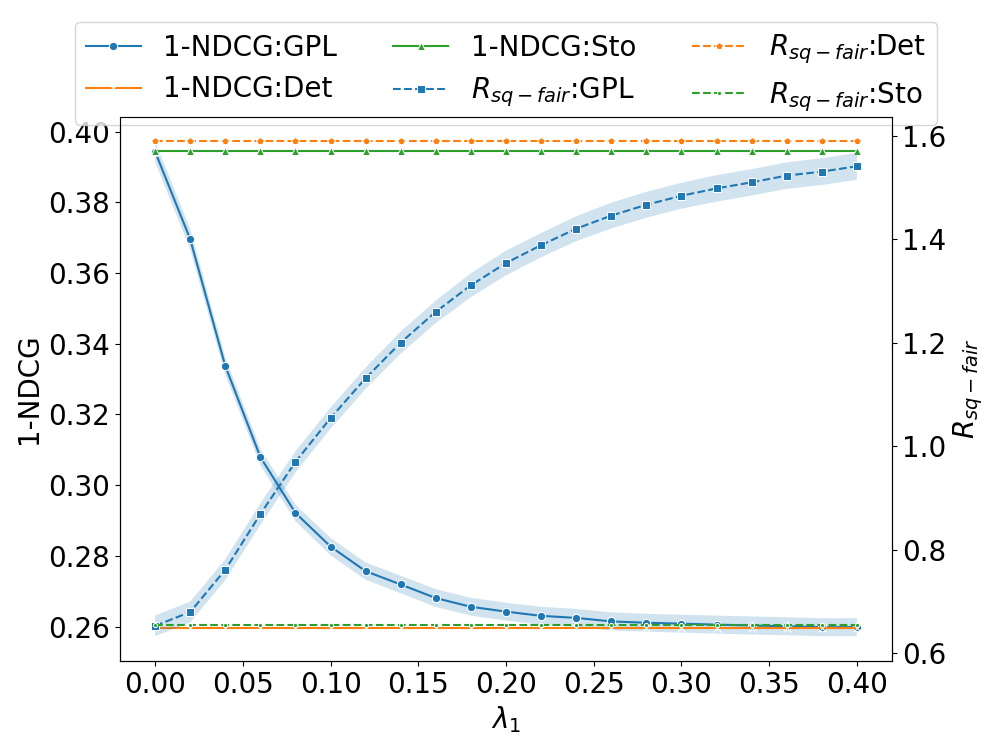}
    \caption{$\zeta=0.9$ with CatBoost on Yahoo.}
  \end{subfigure}%
  \hfill
      \begin{subfigure}{0.33\textwidth}
    \centering
\includegraphics[width=\linewidth]{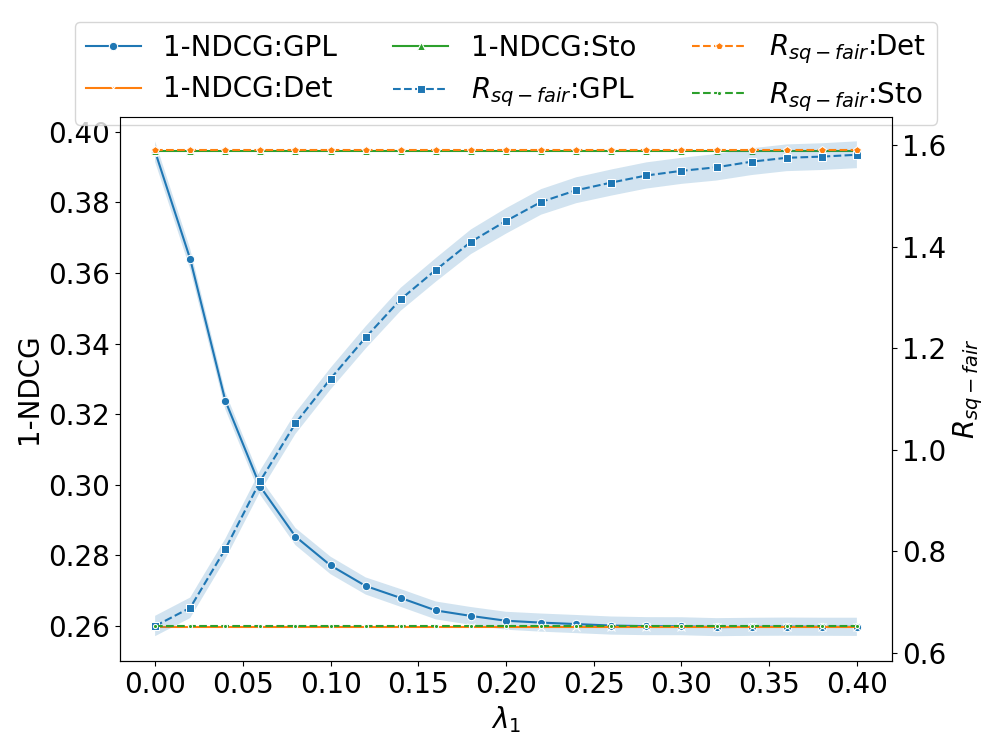}
    \caption{$\zeta=1.0$ with CatBoost on Yahoo.}
  \end{subfigure}%
  \caption{Impact of the scaling factor $\zeta$ on GPL's effectiveness of the trade-off between utility and fairness.
    The x-axis is the threshold $\lambda$, the left (right) y-axis is the risk $R_{{util}}=1-\text{NDCG@5}$ (the disparity $R_{sq-fair}$). Det and Sto are the deterministic and the PL ranking models. The shade is one standard deviation.
    }
  \label{fig:scaling_factor_istella_cat}
\end{figure*}

Fig.~\ref{fig:scaling_factor_istella_cat} shows the utility fairness trade-off results of GPL with varying $\zeta$. As $\zeta$ increases from $0.1$ to $1.0$, when the threshold of the first position $\lambda_1$ increases, the risk $R_{util}$ drops faster and can reach lower minimal value. At the same time, the mean squared disparity $R_{sq-fair}$ increases faster and can end up with larger maximum.
Table~\ref{tab:scaling_factor_yahoo} demonstrates results for coverage and fairness improvement with different values of the scaling factor $\zeta$.
We can observe that, when $\zeta<0.7$, ISRR suffers from large number of abstrintions. This is because the minimum utility risk $R_{util}(\bm{\lambda})$ the model can achieve is not low enough and therefore, is unlikely to find a $\bm{\lambda}$ that results in a UCB $R^{+}_{util}(\bm{\lambda}) < \alpha$ or p-value $p^{HB}(\bm{\lambda}) < \delta$.
When $\zeta=1$, ISRR leads to the least number of absteintions, which implies that it should be adopted if the goal is to provide finite-sample utility guarantee.
However, it has less fairness improvement compared to smaller values of $\zeta$.
%


\begin{table}[H]
\caption{Coverage and FairGain results in Yahoo dataset with thresholds selected by p-value based method, where various values of the scaling factor $\zeta$ are considered.}
\label{tab:scaling_factor_yahoo}
\scriptsize
\begin{tabular}{|ccccccc|}
\hline
\multicolumn{1}{|c|}{}          & \multicolumn{1}{c|}{$\zeta$} & \multicolumn{1}{c|}{Coverage} & \multicolumn{1}{c|}{1-$\alpha$} & \multicolumn{1}{c|}{NDCG@5} & \multicolumn{1}{c|}{FairGain} & \# Abstention \\ \hline
\multicolumn{7}{|c|}{Yahoo}                                                                                                                                                                                                                                  \\ \hline
\multicolumn{1}{|c|}{\multirow{6}{*}{CatBoost+ISRR}} & \multicolumn{1}{c|}{0.1}     & \multicolumn{1}{c|}{-}        & \multicolumn{1}{c|}{-}          & \multicolumn{1}{c|}{-}      & \multicolumn{1}{c|}{-}                                     & 50             \\ \cline{2-7} 
\multicolumn{1}{|c|}{}                          & \multicolumn{1}{c|}{0.3}     & \multicolumn{1}{c|}{-}        & \multicolumn{1}{c|}{-}          & \multicolumn{1}{c|}{-}      & \multicolumn{1}{c|}{-}                                     & 50             \\ \cline{2-7} 
\multicolumn{1}{|c|}{}                          & \multicolumn{1}{c|}{0.5}     & \multicolumn{1}{c|}{100.00\%} & \multicolumn{1}{c|}{0.687}      & \multicolumn{1}{c|}{0.716}  & \multicolumn{1}{c|}{25.50\%}                               & 33             \\ \cline{2-7} 
\multicolumn{1}{|c|}{}                          & \multicolumn{1}{c|}{0.7}     & \multicolumn{1}{c|}{100.00\%} & \multicolumn{1}{c|}{0.687}      & \multicolumn{1}{c|}{0.725}  & \multicolumn{1}{c|}{25.69\%}                               & 4              \\ \cline{2-7} 
\multicolumn{1}{|c|}{}                          & \multicolumn{1}{c|}{0.9}     & \multicolumn{1}{c|}{100.00\%} & \multicolumn{1}{c|}{0.687}      & \multicolumn{1}{c|}{0.726}  & \multicolumn{1}{c|}{25.94\%}                               & 2              \\ \cline{2-7} 
\multicolumn{1}{|c|}{}                          & \multicolumn{1}{c|}{1}       & \multicolumn{1}{c|}{100.00\%} & \multicolumn{1}{c|}{0.687}      & \multicolumn{1}{c|}{0.727}  & \multicolumn{1}{c|}{23.99\%}                               & 2              \\ \hline
\multicolumn{1}{|c|}{\multirow{6}{*}{LightGBM+ISRR}} & \multicolumn{1}{c|}{0.1}     & \multicolumn{1}{c|}{-}        & \multicolumn{1}{c|}{-}          & \multicolumn{1}{c|}{-}      & \multicolumn{1}{c|}{-}                                     & 50             \\ \cline{2-7} 
\multicolumn{1}{|c|}{}                          & \multicolumn{1}{c|}{0.3}     & \multicolumn{1}{c|}{-}        & \multicolumn{1}{c|}{-}          & \multicolumn{1}{c|}{-}      & \multicolumn{1}{c|}{-}                                     & 50             \\ \cline{2-7} 
\multicolumn{1}{|c|}{}                          & \multicolumn{1}{c|}{0.5}     & \multicolumn{1}{c|}{100.00\%} & \multicolumn{1}{c|}{0.687}      & \multicolumn{1}{c|}{0.715}  & \multicolumn{1}{c|}{25.53\%}                               & 35             \\ \cline{2-7} 
\multicolumn{1}{|c|}{}                          & \multicolumn{1}{c|}{0.7}     & \multicolumn{1}{c|}{100.00\%} & \multicolumn{1}{c|}{0.687}      & \multicolumn{1}{c|}{0.725}  & \multicolumn{1}{c|}{25.60\%}                               & 4              \\ \cline{2-7} 
\multicolumn{1}{|c|}{}                          & \multicolumn{1}{c|}{0.9}     & \multicolumn{1}{c|}{100.00\%} & \multicolumn{1}{c|}{0.687}      & \multicolumn{1}{c|}{0.726}  & \multicolumn{1}{c|}{25.85\%}                               & 1              \\ \cline{2-7} 
\multicolumn{1}{|c|}{}                          & \multicolumn{1}{c|}{1}       & \multicolumn{1}{c|}{100.00\%} & \multicolumn{1}{c|}{0.687}      & \multicolumn{1}{c|}{0.727}  & \multicolumn{1}{c|}{20.82\%}                               & 0              \\ \hline
\end{tabular}
\end{table}
 







\vspace{-5pt}
\section{Related Work}


%
\noindent\textbf{Stochastic Ranking and Fairness.}
Stochastic LTR models are initially adopted as it enables optimizing discontinuous ranking metrics (e.g., NDCG@K) using off-the-shelf gradient based methods. A series of work~\citep{wang2018lambdaloss,bruch2020stochastic,ustimenko2020stochasticrank,oosterhuis2018differentiable,oosterhuis2020policy,gao2023policy} shows that scoring functions trained with the PL ranking model can improve rankers' utility in various settings.
%
%
%
More recently, stochastic LTR models are adopted to improve exposure fairness in ranking~\cite{zehlike2022fairness}.
A majority of existing work~\cite{singh2019policy,yadav2021policy,oosterhuis2021computationally,oosterhuis2022learning,gorantla2023optimizing} focuses on developing in-processing methods for training stochastic LTR models towards satisfying a specific notion of fairness in ranking, including group fairness~\cite{singh2019policy}, amortized individual fairness~\cite{biega2018equity,yang2023vertical}, and multiside fairness~\cite{wu2022joint}.
While most existing post-hoc methods fair ranking needs to solve additional optimization problems before inference (e.g., linear programming in~\cite{morik2020controlling}), which is less flexible than our method and cannot offer guaranteed utility or fairness.
%
%
%
In addition, Diaz et al.~\citep{diaz2020evaluating} evaluated stochastic ranking models under different click models in terms of their amortized fairness.
%
%
%
%
Different from existing stochastic LTR models which need training scoring functions from scratch, our method creates a stochastic ranker based on a pre-trained scoring function at inference time.
%
%



\noindent\textbf{Distribution-free Risk Control for Ranking.}
Distribution-free risk control~\cite{bates2021distribution,angelopoulos2022conformal,angelopoulos2021learn,deutschmann2023conformal} provides finite-sample guarantees on the value of a user specified risk function. It is based on split conformal prediction~\citep{papadopoulos2002inductive,angelopoulos2021gentle,lei2014distribution}, a model-agnostic framework for uncertainty quantification and trustworthy prediction.
Bates et al.~\citep{bates2021distribution} propose a method to predict an interval of a pair-wise score of items with guaranteed confidence. It abstains from predicting unconfident pairs. However, it does not directly provide guarantees on list-wise ranking performance.
Angelopoulos et al.~\citep{angelopoulos2022recommendation} apply Learn then Test~\cite{angelopoulos2021learn} to the recall stage of recommendation systems, which creates prediction sets with items having scores higher than a threshold for guaranteed false positive rate.
%
%
%
Wang et al.~\cite{wang2022improving} propose a method based on~\cite{bates2021distribution} to select a threshold for marginal guarantee on the number of candidates from each group and minimizes the prediction set size for each query. It is further extended to the scenario with noisy and biased implicit feedbacks (e.g., clicks)~\cite{wang2022fairness} .
Different from the existing work, this work proposes a more flexible framework to provide finite-sample guarantee on widely used list-wise ranking metrics (e.g., $\text{NDCG@K}$).

\vspace{-8pt}
\section{Conclusion}
In this work, we propose ISRR, an inference-time model-agnostic method for stochastic ranking. It can create a stochastic LTR model with improved amortized individual fairness with a scoring function from a pre-trained LTR model.
With distribution-free risk control, our method can provide guarantee on a user-specified risk function.
%
The integration of the Generalized Plackett-Luce (GPL) model balances utility and fairness.
ISRR avoids expensive training and provides guarantees on a specified metric based on distribution-free risk control. Results on benchmark datasets verify the effectiveness of our proposed method, improving fairness of state-of-the-art deterministic models while ensuring a predefined level of utility.

Despite its promising results, this work is not without limitations.
ISRR can abstain from selecting any threshold from the predefined limited search space in several different situations. First, when subpar pre-trained scoring functions are used. Second, when only calibration sets with small size are available. In addition, using conservative bounds in risk control methods and when $\alpha$ is too small can also make the proposed fail to choose any threshold to achieve finite-sample utility guarantee.
\clearpage
\onecolumn \begin{multicols}{2}
\bibliographystyle{unsrt}
\bibliography{ref}
\end{multicols}
\twocolumn
%
%
%

\clearpage

\appendix

\section{Proof}
\label{sec:proof}

We will use the DKWM inequality to derive a general UCB with discrete risk $R$ because many ranking metrics such as NDCG@K are essentially discrete, which can only take values from a finite set:
\begin{lemma}
    Given a natural number $n$, let $Z_1, Z_2,..., Z_n$ be real-valued independent and identically distributed random variables with cumulative distribution function $F(\cdot)$. Let $F_n$ denote the associated empirical distribution defined by
    \[
    F_n(z) = \frac{1}{n} \sum_{i} 1(Z_i \leq z), ~~z \in \mathbb R
    \]
    Then $\forall \delta \in (0,1)$, with probability at least $1-\delta$, 
    \[
    \sup_{z \in \mathbb R}|F(z) - F_n(z)| \leq \sqrt{\frac{\ln 2/\delta}{2n}}
    \]
    \vspace{-10pt}
    \label{lemma}
\end{lemma}

In order to apply Lemma~\ref{lemma}, we have to show that the risk function $    R^q(\lambda) = \frac{1}{m}\sum_{j=1}^m l(\hat{\by}^j, \by^*)$ can be equivalently written as a function of the empirical cumulative distribution of an indicator random variable w.r.t. a certain threshold $\lambda$.
More specifically, when we take $1-\text{NDCG@K}$ as the risk function $R$ and $l$ as the 0-1 loss function.
Each threshold $\lambda$ corresponds to a certain probability that the best item is shown at the top position, which is denoted as $z(\lambda)$. Denote by $Z_j$ a uniformly random variable defined on $[0,1]$. Then with 0-1 loss $l$
\[
 l(\hat{\by}^j, \by^*) = \mathds{1}(Z_j \leq z(\lambda)),
\]
and therefore the risk function can be rewritten as
\[
R^q(\lambda) =\frac{1}{m}\sum_{j=1}^m \mathds{1}(Z_j \leq z(\lambda)).
\]
The Lemma~\ref{lemma} can therefore apply to bound the difference between the empirical risk $R^q(\lambda)$ and its true expectation $\mathds{E}[R^q(\lambda)]$. Since Lemma~\ref{lemma} covers all $z \in \mathbb R$, it for sure will cover $z(\lambda)$. 

Then we show how to generalize the argument to the case with multiple but discrete loss levels. Suppose the loss values are $l_1, l_2, ... , l_K$ each of which can be the 1-NDCG@K for a query. The chance of incuring a loss $l_k$ is captured by an interval $[z_{k-1}(\lambda), z_k(\lambda))$ with $z_K(\lambda) = 1$:
\[
[0, z_{1}(\lambda)), [z_{1}(\lambda),z_{2}(\lambda)),..., [z_{K-1}(\lambda), 1)
\]
Then 
\[
 l(\hat{\by}^j, \by^*) = \Sigma \cdot \mathds{1}(Z_j \leq z_k(\lambda)) + l_K,
\]
where $\Sigma$ denotes the summation $\sum_{k=1}^{K-1} (l_k - l_{k+1})$.
Then, the empirical CDF can be written as
\[
F_m(z_k(\lambda)) :=\frac{1}{m}\sum_{j=1}^m \mathds{1}(Z_j \leq z_k(\lambda)) 
\]
With this we have the connection between $R^q(\lambda)$ and the empirical CDF as
\begin{align*}
    R^q(\lambda) &= \Sigma \cdot \frac{1}{m}\sum_{j=1}^m \mathds{1}(Z_j \leq z_k(\lambda)) + l_k\\
    &=\Sigma \cdot F_m(z_k(\lambda)) + l_k
\end{align*}
Then we can derive the following inequality
\begin{align*}
    & |  R^q(\lambda) -   \mathds{E}[R^q(\lambda)]| \\ & = \left|(\Sigma \cdot F_m(z_k(\lambda)) + l_k) - (\Sigma \cdot F(z_k(\lambda)) + l_k) \right|\\
    &=\left|\Sigma \cdot (F_m(z_k(\lambda) - F(z_k(\lambda)))\right| \\
    &\leq \sum_{k=1}^{K-1} |l_k - l_{k+1}|\cdot |F_m(z_k(\lambda)) - F(z_k(\lambda))|
\end{align*}
Since for all $z$ we have 
\[
|F_m(z) - F(z)| \leq  \sqrt{\frac{\ln 2/\delta}{2m}}
\]
Then 
\[
  |  R^q(\lambda) -   \mathds{E}[R^q(\lambda)]| \leq \sum_{k=1}^{K-1} |l_k - l_{k+1}| \cdot \sqrt{\frac{\ln 2/\delta}{2m}}, \forall \lambda.
\]
Therefore, we obtained an UCB for the risk $R^q(\lambda)$ when it takes discrete values like 1-NDCG@K. This UCB bounds the difference between the empirical risk $R^q(\lambda)$ computed on calibration data and the true expectation $\mathds{E}[R^q(\lambda)]$.

\section{Experimental Setup Details}
\label{sec:setup_details}


\noindent\textbf{CatBoost~\citep{prokhorenkova2018catboost} and LightGBM~\citep{ke2017lightgbm}.} We perform grid search following the benchmark repository of CatBoost for LTR\footnote{\url{https://github.com/catboost/benchmarks/tree/master/ranking}}. In particular, we search the following hyperparameters: learning rate in $\{0.03, 0.07, 0.15, 0.3\}$, max\_bin in $\{64,128,254\}$, and max\_depth in $\{4,6,8,10\}$. We select the best model is based on NDCG@5 in the validation set.
For the loss function, we follow the aforementioned CatBoost repository to use the ones with the optimal performance in each dataset. For Yahoo, we use \texttt{yeti-rank} for CatBoost and \texttt{RMSE} for LightGBM. For MSLR-WEB30K and Istella-S, we adopt \texttt{QueryRMSE} for CatBoost and \texttt{RMSE} for LightGBM.

\noindent\textbf{Neural Network Scoring Function.} We follow the open-source code\footnote{\url{https://github.com/HarrieO/2022-SIGIR-plackett-luce}} of~\cite{oosterhuis2022learning} to train the NN base model for our method and reproduce results of the state-of-the-art stochastic LTR models.
Specifically, we use a three-layer MLP with sigmoid activation. The number of hidden units is $32$, batch size is $256$. Each model is trained for $100$ epochs. A checkpoint is saved for each epoch. We select the checkpoint with highest NDCG@5 in the validation set.

\section{Complete Experiment Results}
\label{sec:detailed_results}

Here, we demonstrate the complete experimental results.

\subsection{Complete Trade-off Results}
\begin{figure*}[tbh!]
  \centering
  \begin{subfigure}{0.45\textwidth}
    \centering
    \includegraphics[width=\linewidth]{figs/yahoo/interpolation_model_name_cat-query-rmse_scaling_1.0ndcg_sq_fair}
    \caption{CatBoost on Yahoo}
\label{fig:tradeoff_yahoo_cat}
  \end{subfigure}%
  \hfill
  \begin{subfigure}{0.45\textwidth}
    \centering
    \includegraphics[width=\linewidth]{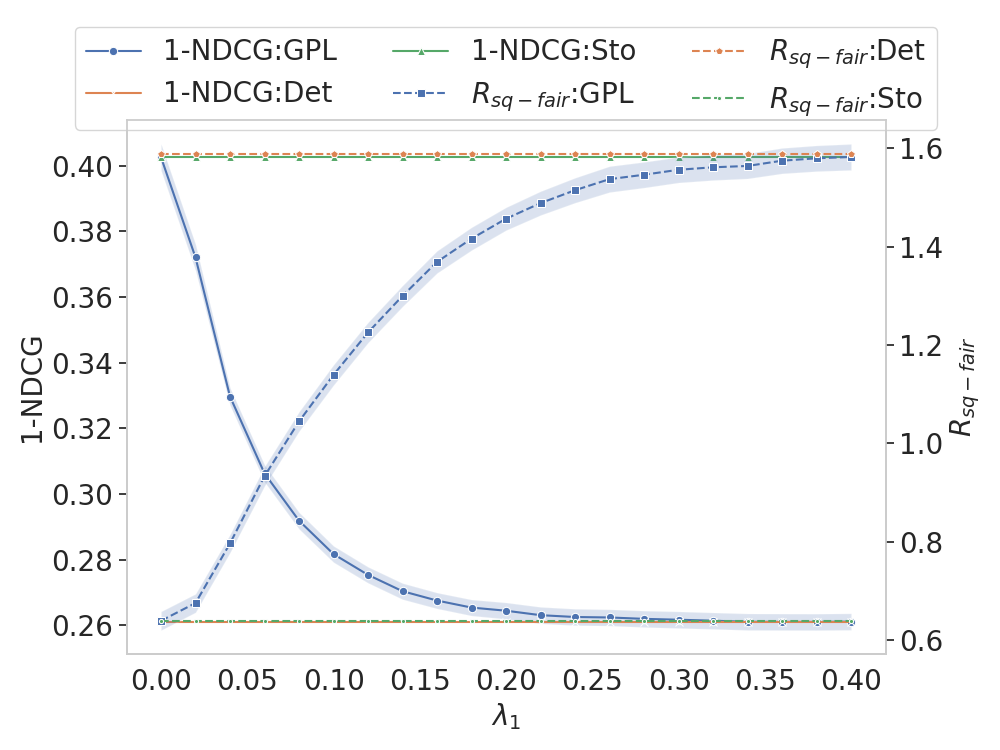}
    \caption{LGB on Yahoo}
\label{fig:tradeoff_yahoo_cat}
  \end{subfigure}%
  \hfill
  \begin{subfigure}{0.45\textwidth}
    \centering
\includegraphics[width=\linewidth]{./figs/yahoo/interpolation_model_name_lambdaloss_scaling_1.0ndcg_sq_fair.png}
    \caption{NN on Yahoo}%
  \end{subfigure}%
  \hfill
  \begin{subfigure}{0.45\textwidth}
  \includegraphics[width=\linewidth]{./figs/mslr/interpolation_model_name_cat-query-rmse_scaling_1.0ndcg_sq_fair.png}
    \caption{CatBoost on MSLR-WEB30K}
    \label{fig:tradeoff_mslr_cat}
  \end{subfigure}%
  \hfill
   \begin{subfigure}{0.45\textwidth}
  \includegraphics[width=\linewidth]{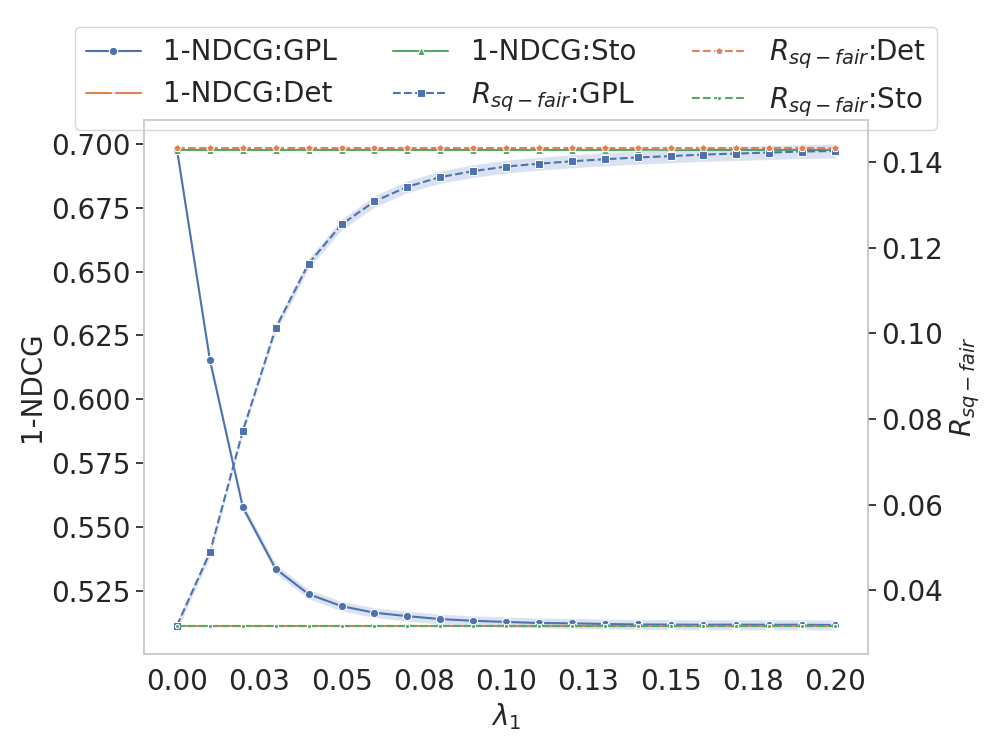}
    \caption{CatBoost on MSLR-WEB30K}
    \label{fig:tradeoff_mslr_cat}
  \end{subfigure}%
  \hfill
  \begin{subfigure}{0.45\textwidth}
    \centering
\includegraphics[width=\linewidth]{./figs/mslr/interpolation_model_name_lambdaloss_scaling_1.0ndcg_sq_fair}
    \caption{NN on MSLR-WEB30K}
\label{fig:tradeoff_mslr_NN}
  \end{subfigure}%
  \caption{Complete results for trade-off between utility and fairness achieved by GPL.
    The x-axis is the threshold $\lambda$, the left (right) y-axis is the risk $R_{{util}}=1-\text{NDCG@5}$ (the disparity $R_{sq-fair}$). Det and Sto are the deterministic and the PL ranking models. The shade is one standard deviation. As $\lambda$ increases, GPL only includes items with higher scores, leading to lower risk and higher disparity.}
  \label{fig:tradeoff_full}
\end{figure*}

Here, Fig.~\ref{fig:tradeoff_full} shows the complete results of utility fairness trade-off for the two datasets with all three scoring functions pretrained with deterministic LTR models.

\begin{table}[tbh!]
\caption{ Coverage and FairGain results with thresholds $\lambda$ selected by DKMW inequality.}
\small
\begin{tabular}{|c|ccccc|}
\hline
         & \multicolumn{5}{c|}{Yahoo}                                                                                                                                              \\ \hline
         & \multicolumn{1}{c|}{Coverage} & \multicolumn{1}{c|}{1-$\alpha$} & \multicolumn{1}{c|}{NDCG@5} & \multicolumn{1}{c|}{FairGain} & \# Abstain \\ \hline
CatBoost+ISRR & \multicolumn{1}{c|}{100\%}    & \multicolumn{1}{c|}{0.687}      & \multicolumn{1}{c|}{0.734}  & \multicolumn{1}{c|}{13.69\%}                               & 16         \\ \hline
LGB+ISRR     & \multicolumn{1}{c|}{100\%}    & \multicolumn{1}{c|}{0.687}      & \multicolumn{1}{c|}{0.734}  & \multicolumn{1}{c|}{10.95\%}                               & 16         \\ \hline
NN+ISRR      & \multicolumn{1}{c|}{100\%}    & \multicolumn{1}{c|}{0.641}      & \multicolumn{1}{c|}{0.682}  & \multicolumn{1}{c|}{30.37\%}                               & 0          \\ \hline
         & \multicolumn{5}{c|}{MSLR}                                                                                                                                               \\ \hline
CatBoost & \multicolumn{1}{c|}{100\%}    & \multicolumn{1}{c|}{0.449}      & \multicolumn{1}{c|}{0.484}  & \multicolumn{1}{c|}{11.19\%}                               & 12         \\ \hline
LGB      & \multicolumn{1}{c|}{100\%}    & \multicolumn{1}{c|}{0.449}      & \multicolumn{1}{c|}{0.483}  & \multicolumn{1}{c|}{8.39\%}                                & 15         \\ \hline
NN       & \multicolumn{1}{c|}{100\%}    & \multicolumn{1}{c|}{0.405}      & \multicolumn{1}{c|}{0.435}  & \multicolumn{1}{c|}{22.22\%}                               & 0          \\ \hline
\end{tabular}
\end{table}

\subsection{Discussion on Running Time Results of \cite{oosterhuis2021computationally}}
\label{subsec:issue_of_pl_rank}

As shown in the publicly available reporitories\footnote{\url{https://github.com/HarrieO/2021-SIGIR-plackett-luce/blob/master/algorithms/lambdaloss.py}}\footnote{\url{https://github.com/HarrieO/2022-SIGIR-plackett-luce/blob/main/algorithms/lambdaloss.py}} from~\cite{oosterhuis2021computationally,oosterhuis2022learning}, we can observe that the implementation of LambdaLoss is the deterministic version. So, it is not necessary to use multiple sampled rankings for the deterministic method LambdaLoss. This can also be verified by the results shown in Fig. 1 of~\cite{oosterhuis2021computationally}, when the number of sampled rankings increases, the NDCG@K of LambdaLoss does not change.

\subsection{Additional Results with DKWM Inequality}
Here, we show the coverage rates by DKWM inequality. We can observe that, compared to Hoeffding-Benktus, DKWM inequality leads to a less tight UCB. This explains why DKWM results in higher NDCG@5, less drop in the disparity measure $R_{sq-fair}$ and larger number of absteintions. We leave finding a tighter bound than the Hoeffding-Benktus to future work.

\begin{figure*}[tbh!]
  \centering
  
  \begin{subfigure}{0.4\textwidth}
    \centering
    \includegraphics[width=\linewidth]{figs/yahoo/HB/cat-query-rmse_scaling_1.0_alpha_0.3133_delta_0.05_cal_ratio_0.25_07_51_55.png}
\subcaption{\cat \ on Yahoo}
    \label{fig:coverage_cat_yahoo}
  \end{subfigure}%
  \hfill
    \begin{subfigure}{0.4\textwidth}
    \centering
    \includegraphics[width=\linewidth]{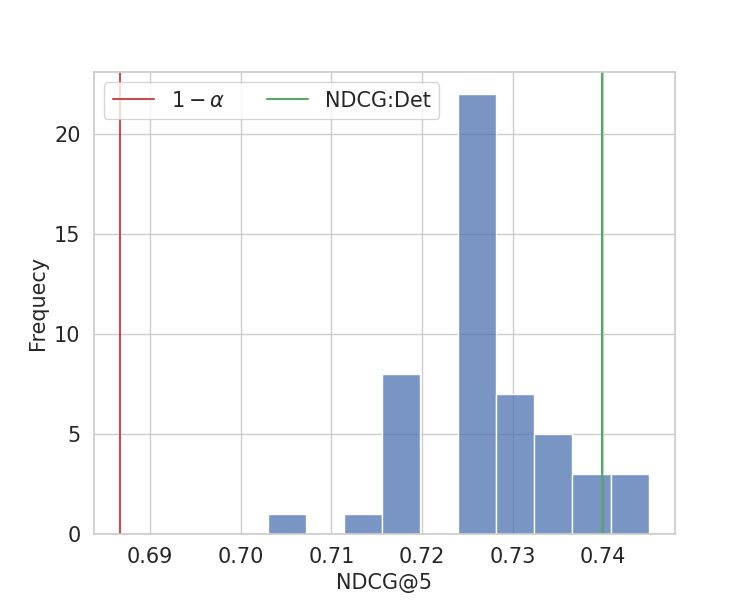}
\subcaption{LGB \ on Yahoo}
    \label{fig:coverage_cat_yahoo}
  \end{subfigure}%
  \hfill
  \begin{subfigure}{0.4\textwidth}
    \centering
    \includegraphics[width=\linewidth]{figs/yahoo/HB/lambdaloss_scaling_1.0_alpha_0.35919999999999996_delta_0.05_cal_ratio_0.25_07_49_27}
    \caption{NN on Yahoo}
\label{fig:coverage_NN_yahoo}
  \end{subfigure}%
\hfill
  \begin{subfigure}{0.4\textwidth}
    \centering
    \includegraphics[width=\linewidth]{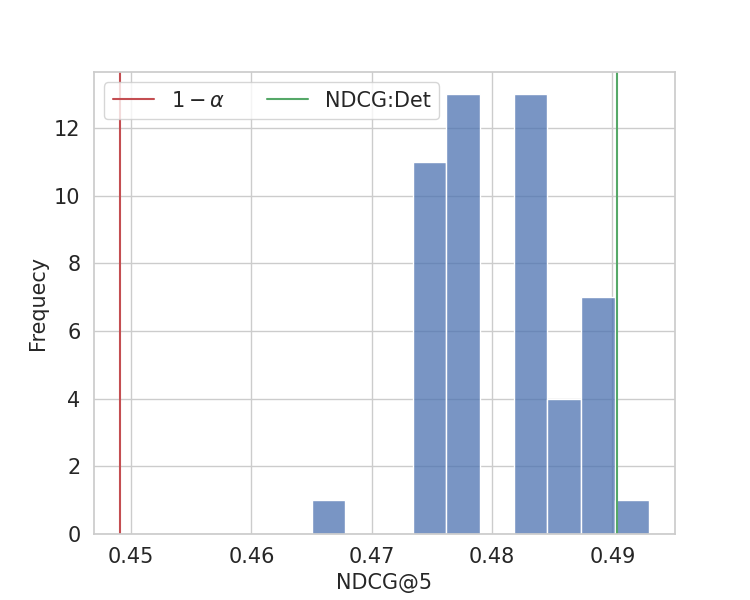}
\subcaption{\cat \ on MSLR}
    \label{fig:coverage_cat_yahoo}
  \end{subfigure}%
  \hfill
    \begin{subfigure}{0.4\textwidth}
    \centering
    \includegraphics[width=\linewidth]{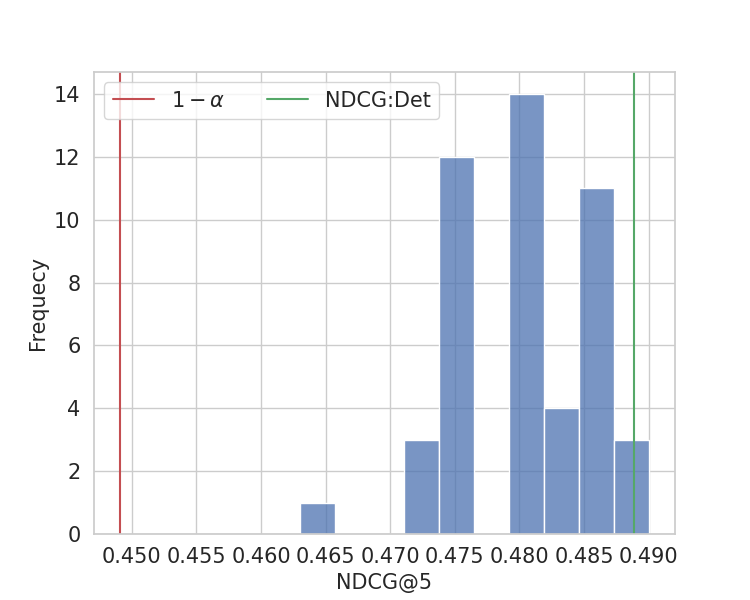}
\subcaption{LGB \ on MSLR}
    \label{fig:coverage_cat_yahoo}
  \end{subfigure}%
  \hfill
  \begin{subfigure}{0.4\textwidth}
    \centering
    \includegraphics[width=\linewidth]{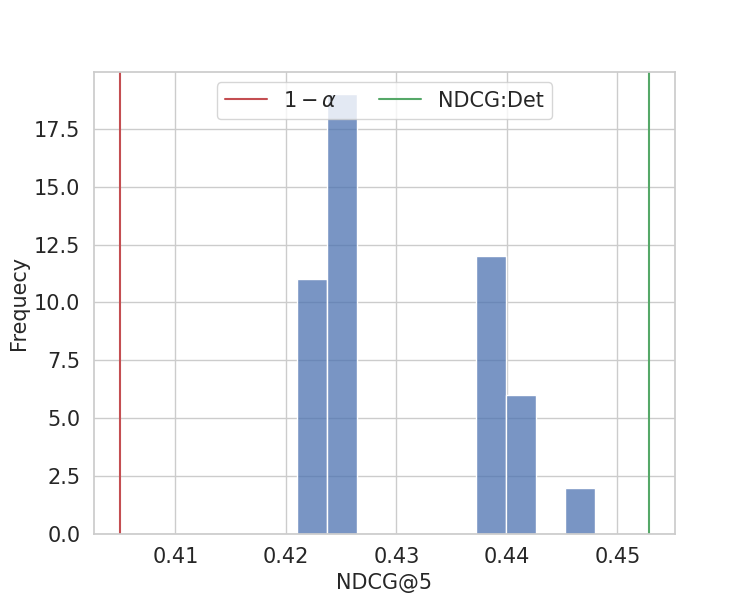}
    \caption{NN on MSLR}
\label{fig:coverage_NN_yahoo}
  \end{subfigure}%
  \caption{Distribution of NDCG@5 achieved by \ourmodel over $50$ runs with $\lambda$ selected by Hoeffding-Benktus. 
  The red (green) vertical line is the desired NDCG@5 level $1-\alpha = 0.9 U^*$. $U^*$ is the NDCG@5 of the deterministic model.}
  \label{fig:coverage}
\end{figure*}

Then, for the two datasets, we show more detailed results on coverage -- the distributions of NDCG@5 with all three scoring functions in Fig.~\ref{fig:coverage}.
We make the following observations.
\begin{itemize}
    \item With thresholds selected by both Hoeffing-Benktus and DKWM, our method can achieve 100\% coverage rate to provide the guarantee that NDCG@5$\ge 1- \alpha$ on both datasets with all three scoring functions.
    \item The UCB of DKWM is less tight than that of Hoeffing-Benktus, leading to more conservative selection of thresholds, higher NDCG@5, worse improvement in fairness, and greater number of absteintions.
\end{itemize}

\begin{figure*}[tbh!]
  \centering
  
  \begin{subfigure}{0.4\textwidth}
    \centering
    \includegraphics[width=\linewidth]{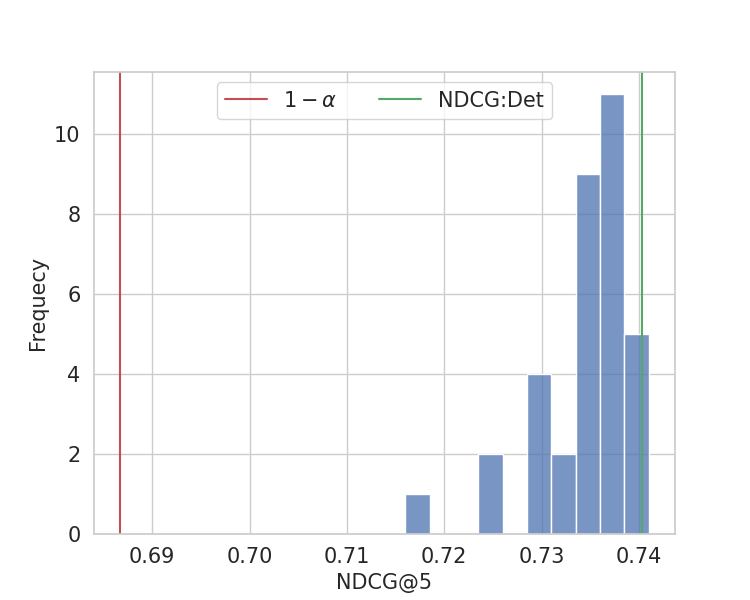}
\subcaption{\cat \ on Yahoo}
    \label{fig:coverage_cat_yahoo}
  \end{subfigure}%
  \hfill
    \begin{subfigure}{0.4\textwidth}
    \centering
    \includegraphics[width=\linewidth]{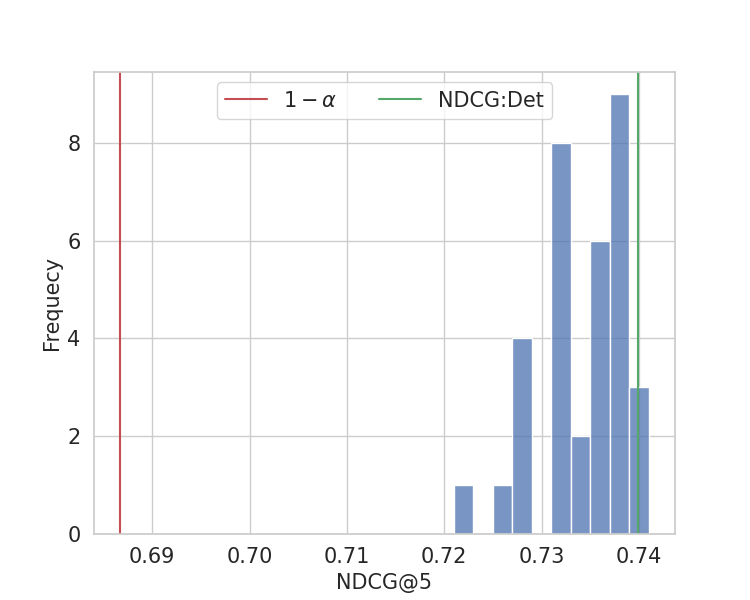}
\subcaption{LGB \ on Yahoo}
    \label{fig:coverage_cat_yahoo}
  \end{subfigure}%
  \hfill
  \begin{subfigure}{0.4\textwidth}
    \centering
    \includegraphics[width=\linewidth]{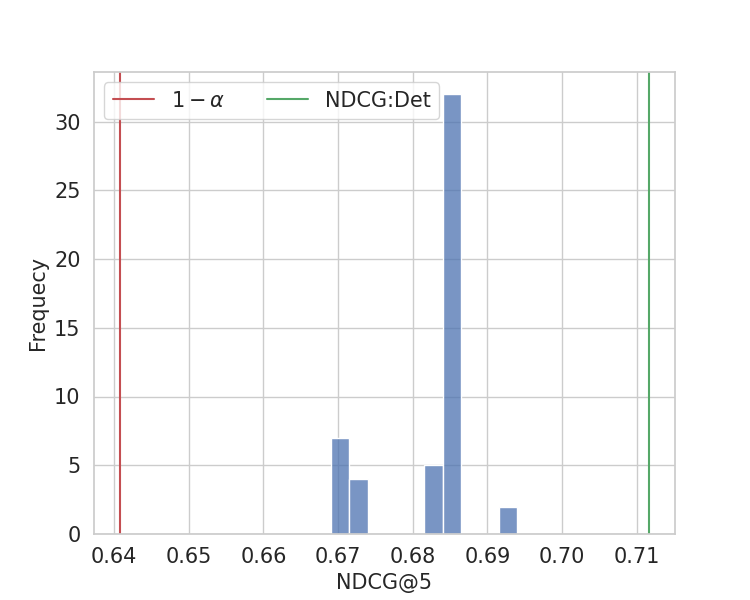}
    \caption{NN on Yahoo}
\label{fig:coverage_NN_yahoo}
  \end{subfigure}%
\hfill
  \begin{subfigure}{0.4\textwidth}
    \centering
    \includegraphics[width=\linewidth]{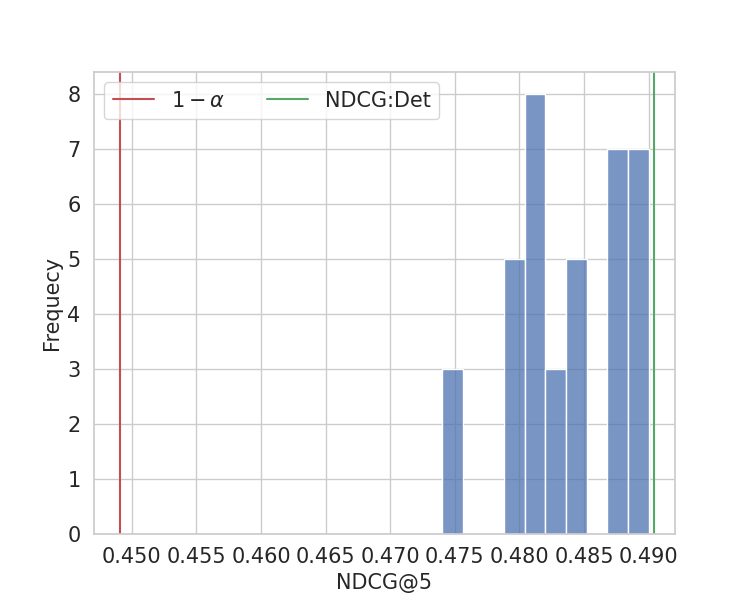}
\subcaption{CatBoost on MSLR}
    \label{fig:coverage_cat_yahoo}
  \end{subfigure}%
  \hfill
    \begin{subfigure}{0.4\textwidth}
    \centering
    \includegraphics[width=\linewidth]{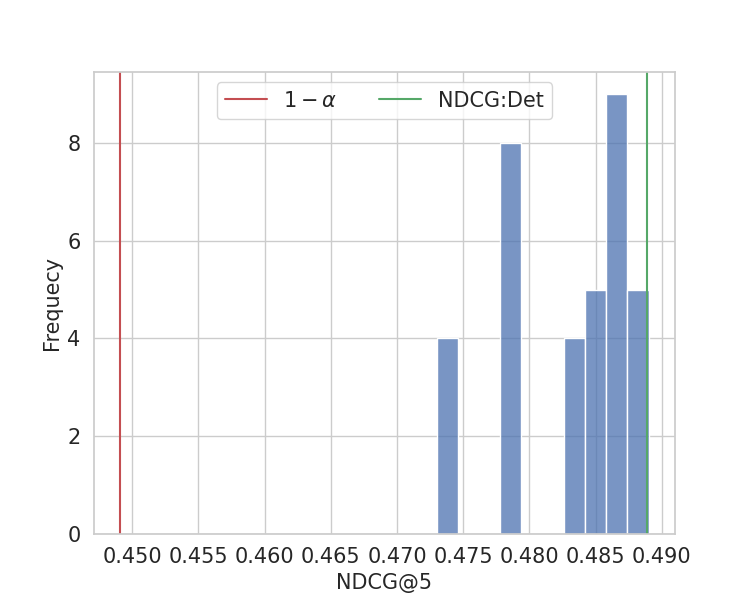}
\subcaption{LGB \ on MSLR}
    \label{fig:coverage_cat_yahoo}
  \end{subfigure}%
  \hfill
  \begin{subfigure}{0.4\textwidth}
    \centering
    \includegraphics[width=\linewidth]{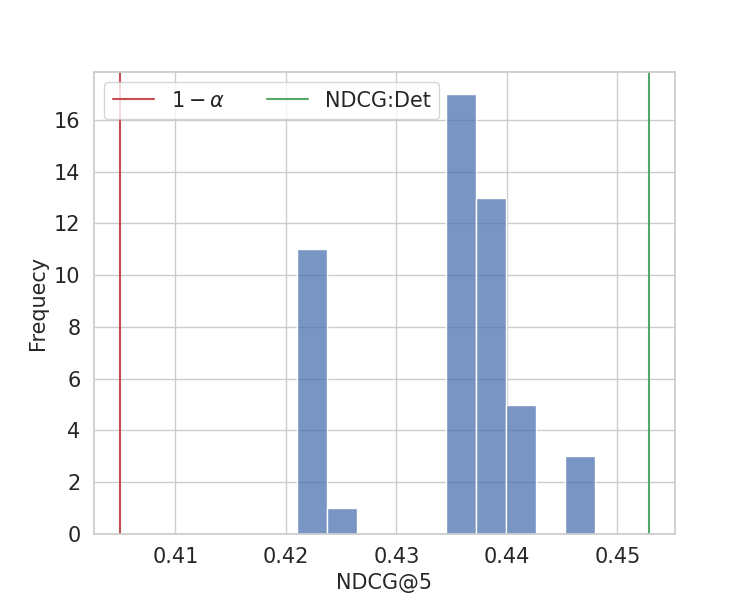}
    \caption{NN on MSLR}
\label{fig:coverage_NN_yahoo}
  \end{subfigure}%
  \caption{Distribution of NDCG@5 achieved by ISRR over $50$ runs with $\lambda$ selected by DKWM inequality. 
  The red (green) vertical line is the desired NDCG@5 level $1-\alpha = 0.9 U^*$. $U^*$ is the NDCG@5 of the deterministic model.
  }
  \label{fig:coverage_DKWM}
\end{figure*}


\subsection{Coverage and FairGain Results for Stochastic LTR models}
\label{subsec:compare_w_sto}
Here, although it is undesirable to apply our methods on the top of pre-trained stochastic LTR models due to their training cost, we still illustrate the complete coverage and fairness gain results for the state-of-the-art stochastic LTR models including PL-Rank-3~\cite{oosterhuis2022learning}, StochasticRank~\cite{ustimenko2020stochasticrank}, and Policy Gradient~\cite{singh2019policy} along with CatBoost, LightGBM, and LambdaLoss.
With the results shown in Table~\ref{tab:complete_coverage_results}, we make the following observations:
\begin{itemize}
    \item First, the proposed method achieves effective utility fairness trade-off while maintaining the guarantee on utility. With the scoring function from the pretrained stochastic LTR models, our method can improve amortized individual fairness by at least 30.52\% on Yahoo and 26.14\% on MSLR, respectively, with guarantee that the NDCG@5 is at least 90\% of that obtained with the same scoring function on the top of a deterministic ranking model.
    \item Second, with PL-Rank-3, our method achieves the most effective trade-off with 35.05\% and 29.61\% improvement in amortized individual fairness measured by $R_{sq-fair}$. We conjecture that this is due to the scoring function pretrained by PL-Rank-3 maintains the best ranking performance.
\end{itemize}

\begin{table*}[tbh!]
\caption{Coverage and fairness gain results with thresholds $\lambda$ selected by HB inequality. All pre-trained scoring functions from both deterministic and stochastic LTR models are optimized for utlity.}
\label{tab:complete_coverage_results}
\begin{tabular}{|cccccc|}
\hline
\multicolumn{6}{|c|}{Yahoo}                                                                                                                                                                                     \\ \hline
\multicolumn{1}{|c|}{}                & \multicolumn{1}{c|}{Coverage} & \multicolumn{1}{c|}{1-$\alpha$} & \multicolumn{1}{c|}{NDCG@5} & \multicolumn{1}{c|}{FairGain} & \# Abstain \\ \hline
\multicolumn{1}{|c|}{CatBoost+ISRR}        & \multicolumn{1}{c|}{100\%}    & \multicolumn{1}{c|}{0.687}      & \multicolumn{1}{c|}{0.727}  & \multicolumn{1}{c|}{23.99\%}                               & 2          \\ \hline
\multicolumn{1}{|c|}{LightGBM+ISRR}        & \multicolumn{1}{c|}{100\%}    & \multicolumn{1}{c|}{0.687}      & \multicolumn{1}{c|}{0.727}  & \multicolumn{1}{c|}{20.77\%}                               & 0          \\ \hline
\multicolumn{1}{|c|}{LambdaLoss+ISRR}      & \multicolumn{1}{c|}{100\%}    & \multicolumn{1}{c|}{0.641}      & \multicolumn{1}{c|}{0.673}  & \multicolumn{1}{c|}{35.05\%}                               & 0          \\ \hline
\multicolumn{1}{|c|}{PL-Rank-3+ISRR}       & \multicolumn{1}{c|}{100\%}    & \multicolumn{1}{c|}{0.641}      & \multicolumn{1}{c|}{0.673}  & \multicolumn{1}{c|}{35.05\%}                               & 0          \\ \hline
\multicolumn{1}{|c|}{Policy Gradient+ISRR} & \multicolumn{1}{c|}{100\%}    & \multicolumn{1}{c|}{0.641}      & \multicolumn{1}{c|}{0.674}  & \multicolumn{1}{c|}{30.52\%}                               & 0          \\ \hline
\multicolumn{1}{|c|}{StochasticRank+ISRR}  & \multicolumn{1}{c|}{100\%}    & \multicolumn{1}{c|}{0.641}      & \multicolumn{1}{c|}{0.673}  & \multicolumn{1}{c|}{34.88\%}                               & 0          \\ \hline
\multicolumn{6}{|c|}{MSLR}                                                                                                                                                                                      \\ \hline
\multicolumn{1}{|c|}{CatBoost+ISRR}        & \multicolumn{1}{c|}{100\%}    & \multicolumn{1}{c|}{0.449}      & \multicolumn{1}{c|}{0.481}  & \multicolumn{1}{c|}{16.08\%}                               & 0          \\ \hline
\multicolumn{1}{|c|}{LightGBM+ISRR}        & \multicolumn{1}{c|}{100\%}    & \multicolumn{1}{c|}{0.449}      & \multicolumn{1}{c|}{0.480}  & \multicolumn{1}{c|}{13.29\%}                               & 2          \\ \hline
\multicolumn{1}{|c|}{LambdaLoss+ISRR}      & \multicolumn{1}{c|}{100\%}    & \multicolumn{1}{c|}{0.405}      & \multicolumn{1}{c|}{0.430}  & \multicolumn{1}{c|}{26.14\%}                               & 0          \\ \hline
\multicolumn{1}{|c|}{PL-Rank-3+ISRR}       & \multicolumn{1}{c|}{100\%}    & \multicolumn{1}{c|}{0.405}      & \multicolumn{1}{c|}{0.430}  & \multicolumn{1}{c|}{29.61\%}                               & 0          \\ \hline
\multicolumn{1}{|c|}{Policy Gradient+ISRR} & \multicolumn{1}{c|}{100\%}    & \multicolumn{1}{c|}{0.405}      & \multicolumn{1}{c|}{0.430}  & \multicolumn{1}{c|}{26.14\%}                               & 0          \\ \hline
\multicolumn{1}{|c|}{StochasticRank+ISRR}  & \multicolumn{1}{c|}{100\%}    & \multicolumn{1}{c|}{0.405}      & \multicolumn{1}{c|}{0.429}  & \multicolumn{1}{c|}{27.63\%}                               & 0          \\ \hline
\end{tabular}
\end{table*}

\section{Score for Distribution-free Risk Control}
\label{sec:rc_score}
%
%
Risk control (RC) score is a crucial design choice for distribution-free risk control~\citep{bates2021distribution,angelopoulos2021learn}.
In GPL, RC score determines whether an item would be included in the prediction set $\mathcal{T}(\lambda_k)$ for position $k$.
%
%
%
A natural choice of RC score is the expected position-wise sampling probability $\mathbb{E}_{\by_{1:k-1}}[p(d|\by_{1:k-1},0)]$ in the PL model, which is a calibrated measure of the predicted relevance of item $d$ for position $k$.
However, it can be computationally expensive to compute $p(d|\by_{1:k-1},0)$ for all possible $(d,\by_{1:k-1})$ when $|\mathcal{D}^q|$ and $k$ are large.
%
%
To decouple the score function of position $k$ from the sampled partial rankings $\by_{1:k-1}$, we propose to use the probability to sample $d$ at the first position in the PL model, i.e., $p(d|\emptyset,0)$, as the RC score.
Unlike the original ranking score $s_d$, as probability density, $p(d|\emptyset,0) \in [0,P^{max}]$ for $d\in \mathcal{D}^q$ for all queries $q$, is always bounded. In addition, the maximum $P^{max} = \max_d(p(d|\emptyset,0))$ can be easily computed using the calibration set, making it convenient to obtain the range for the threhsolds $\lambda_k$ for creating the set of candidate thresholds $\hat{\Lambda}$ for distribution-free risk control.
%
%
%
Furthermore, it can be easily demonstrated that $p(d|\emptyset,0)$ exhibits monotonicity with respect to  the expected position-wise sampling probability $\mathbb{E}_{\by{1:k-1}}[p(d|\by_{1:k-1},0)]$.

%

%

%

In practice, we use the softmax of normalized ranking scores from pretrained scoring function $f$ as the scores:
\begin{equation}
\begin{split}
   s_d  & = (s_d^{raw} - mean({s}^{raw})) / std(s^{raw}) \\
    p_d & = softmax(s_d), \\
\end{split}
\end{equation}
where $mean({s}^{raw})$ and $std(s^{raw})$ are the mean and standard deviation of the ranking scores computed on the validation set.
%

Empirically, we find it is much easier to design the range for searching $\lambda$ if the scores are normalized.
Recall that ranking scores are unbounded real values, normalizing them makes the range of scores, i.e., $p(d|\emptyset)$. much easier to compute.

\end{document}